\pdfoutput=1

\documentclass[dvipsnames,11pt]{article}

\usepackage{ACL2023}

\usepackage{times}
\usepackage{latexsym}

\usepackage[T1]{fontenc}

\usepackage[utf8]{inputenc}

\usepackage{microtype}

\usepackage{inconsolata}

\usepackage{booktabs}
\usepackage{graphicx}
\usepackage{caption}
\usepackage{subcaption}
\usepackage{algorithm}
\usepackage{algpseudocode}
\usepackage{multirow}
\usepackage{tabularx}
\usepackage{amsmath}
\usepackage{amssymb}
\usepackage{pifont}

\newcommand{\parlaistandard}{\textsc{Standard}}
\newcommand{\parlaiadversarial}{\textsc{Adversarial}}
\newcommand{\gbair}{\textbf{G-BAIR}}
\newcommand{\random}{\textbf{Random}}
\newcommand{\sentencetfive}{\textbf{SentenceT5}}
\newcommand{\use}{\textbf{USE}}
\newcommand{\tfivebase}{\textsc{T5 Base}}
\newcommand{\tfivexxl}{\textsc{T5 XXL}}
\newcommand{\palm}{\textsc{PaLM 62B}}

\title{Gradient-Based Automated Iterative Recovery for\\Parameter-Efficient Tuning}

\author{Maximilian Mozes$^{1,2}$\thanks{\quad Work done during an internship at Google Research.}\qquad Tolga Bolukbasi$^1$\qquad Ann Yuan$^1$\qquad Frederick Liu$^1$\\\textbf{Nithum Thain$^1$\qquad Lucas Dixon$^1$}\\
$^1$Google Research\\
$^2$University College London\\
\small{\{\texttt{tolgab,annyuan,frederickliu,nthain,ldixon\}@google.com}}\\
\small{\texttt{maximilian.mozes@ucl.ac.uk}}
}

\begin{document}
\maketitle
\begin{abstract}
Pretrained large language models (LLMs) are able to solve a wide variety of tasks through transfer learning. 
Various explainability methods have been developed to investigate their decision making process. TracIn~\cite{pruthi2020estimating} is one such gradient-based method which explains model inferences based on the influence of  training examples. 
In this paper, we explore the use of TracIn to improve model performance in the parameter-efficient tuning (PET) setting. 
We develop conversational safety classifiers via the prompt-tuning PET method and show how the unique characteristics of the PET regime enable TracIn to identify the cause for certain misclassifications by LLMs. 
We develop a new methodology for using gradient-based explainability techniques to improve model performance, \gbair{}: \textit{gradient-based automated iterative recovery}. 
We show that \gbair{} can recover LLM performance on benchmarks after manually corrupting training labels.
This suggests that influence methods like TracIn can be used to automatically perform data cleaning, and introduces the potential for interactive debugging and relabeling for PET-based transfer learning methods.
\end{abstract}

\section{Introduction}

\begin{figure}[!ht]
    \centering
    \includegraphics[width=\columnwidth]{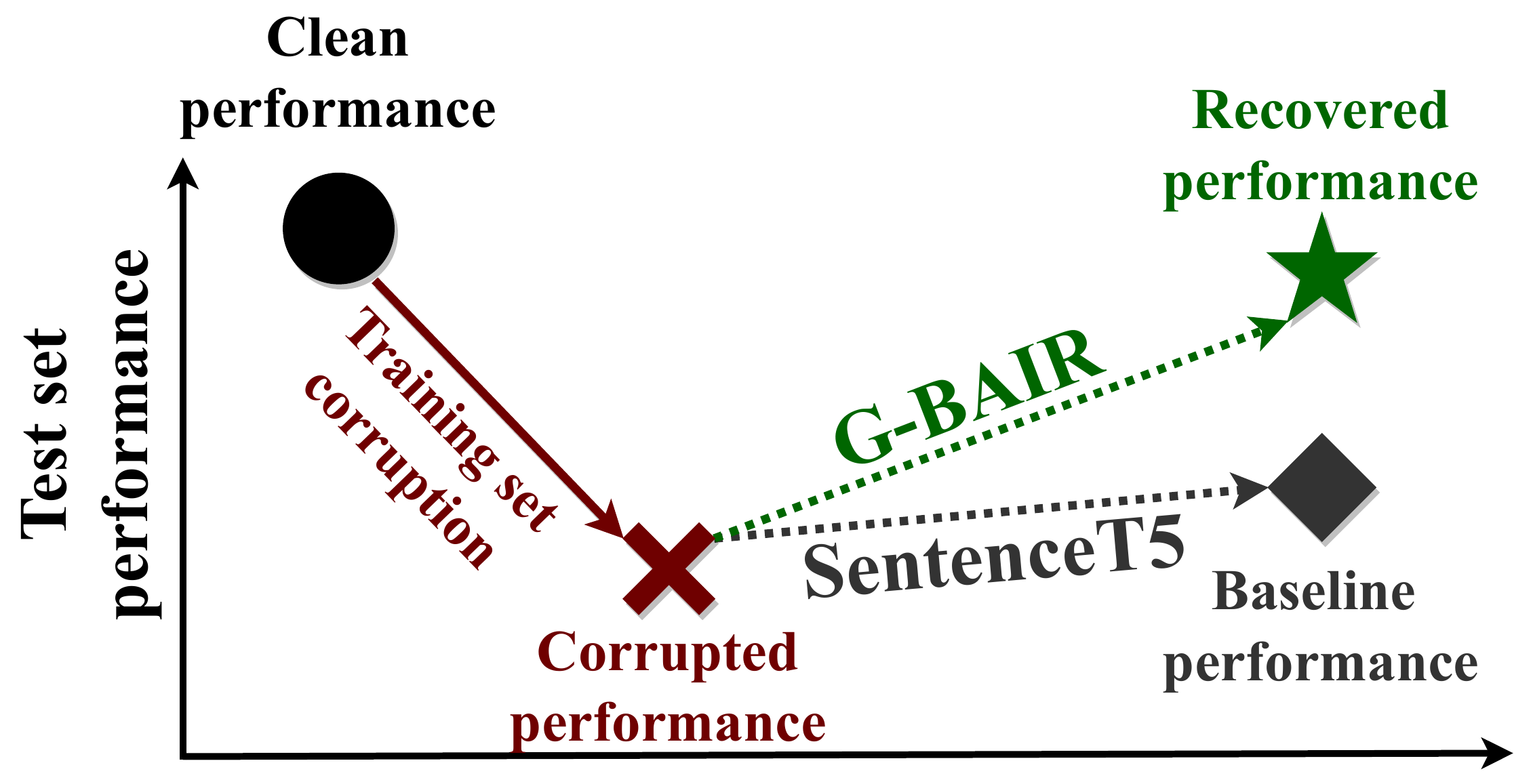}
    \caption{Illustration of our \gbair{} method used to recover prompt-tuning model performance drops incurred through data corruption. Clean model performance ({\color{Black}{\ding{108}}}) drops as a result of training data corruption ({\color{Sepia}{\ding{54}}}). \gbair{} ({\color{OliveGreen}$\bigstar$}) can be applied to identify and mitigate corrupted examples, thereby recovering clean test set performance better than the compared \sentencetfive{} ({\color{darkgray}{\ding{117}}}) baseline.}
    \label{fig:tracin_tldr_plot}
\end{figure}

Pretrained large language models (LLMs) are Transformer-based models~\cite{vaswani2017attention} with hundreds of millions, or even billions of parameters trained on large datasets containing hundreds of billions of words~\cite{raffel2020exploring, brown2020language,chowdhery2022palm}. 
LLMs have recently become ubiquitous due to their ability to solve a wide range of problems, and their capacity for transfer learning with relatively little data. 
Researchers have explored three approaches for transfer learning: (1) in-context few shot learning which requires only a handful of examples~\cite{radford2019language,brown2020language,schick2021s}, (2) fine-tuning the entire model on large datasets containing thousands of examples~\cite{peters-etal-2018-deep,Devlin2019BERTPO}, and (3) parameter-efficient tuning (PET), in which only a small number of model parameters (e.g., a few thousand) are tuned~\cite{li2021prefix,liu2022few}.
This last approach has been shown to outperform in-context learning and achieve comparable performance to fine-tuning given only moderately sized datasets containing hundreds of examples~\cite{agrawal2022qameleon}. 

An advantage of using smaller datasets and training fewer parameters is that it becomes possible to iteratively improve the resulting model, for example upon observing incorrect predictions on a test set. 
To do so requires interpreting the underlying cause of incorrect predictions.
Various techniques have been developed for this purpose. 
Popular approaches include saliency methods, like integrated gradients or \textit{SHapley Additive exPlanations}~\cite[SHAP;][]{shapley} which identify key features that the model is using in its calculation, and training data attribution methods, like TracIn~\cite{pruthi2020estimating} and influence functions~\cite{koh2017understanding} which retrieve the most relevant training examples based on their influence on a test prediction. 
Beyond explainability, these techniques have also been applied for mitigation to improve model performance, by manipulating either {highlighted} features or training examples.

In this paper, we demonstrate the efficacy of TracIn for parameter-efficient tuning. 
This recipe has a number of unique advantages. 
Using TracIn with whole model fine-tuning is intractable without approximation techniques, like layer selection or gradient projection~\cite{yeh2022first}, due to in-practice memory constraints. 
By contrast with PET, we are working with both a smaller training dataset and a smaller number of training parameters.
Thus when using TracIn with PET, we are able to compute the exact influence of each training example on a test prediction.

We introduce the \textit{Gradient-Based Automated Iterative Recovery} (\gbair{}) protocol, by which we iteratively improve a PET model through identifying examples using TracIn that are responsible for lowering model performance in a validation set (Figure~\ref{fig:tracin_tldr_plot}). 
We develop a corrupted data benchmark on two datasets related to offensive content and toxicity detection, \textsc{ParlAI Single} \parlaistandard{} and \textsc{ParlAI Single} \parlaiadversarial{}~\cite{dinan-etal-2019-build}, to evaluate our protocol for identifying mislabeled examples and improving model performance. 
Using the recently proposed \tfivebase{}, \tfivexxl{}~\cite{raffel2020exploring}, and \palm{}~\cite{chowdhery2022palm} LLMs, we show that our protocol is able to recover a significant portion of the precision that is lost by corrupted data labels for both datasets, thereby outperforming both random and semantics-based baselines.

\section{Related work}

\paragraph{Parameter-efficient tuning and data quality}
Recently, methods for parameter-efficient tuning of language models have been introduced that are effective with smaller datasets~\cite[i.e., 100s of examples;][]{liu2022few,agrawal2022qameleon}.
However, commonly used natural language processing (NLP) datasets, including benchmark sets, have been discovered to contain noise in the form of typos, spelling mistakes, and even mislabelings~\cite{kumar2020noisy, northcutt2021errors}.
The smaller the dataset, the higher the price of mislabeled examples.
Automated dataset denoising techniques have been proposed~\cite{muller2019mislabeling}. Although one can employ multiple strategies to achieve cleaner datasets, our goal is to identify the noisy examples that actually affect the model predictions. We take a model interpretability-based approach and choose to ignore examples that appear to have no effect on the model quality. This is different from standard data cleaning approaches where the focus is on final dataset quality independent of the model.

\paragraph{Influence functions and other applications of TracIn}

Earlier methods for studying the influence of training examples on model parameters scale poorly with the number of model parameters and dataset size~\cite{koh2017understanding,Yeh2018RepresenterPS}. More recent methods address some of the scalability issues of the original influence functions through approximations~\cite{schioppa2021scaling}.~\citet{basu2021influence} show that the original formulation is fairly accurate for shallow networks, but are often noisy for deeper networks. In this paper, we focus on TracIn~\cite{pruthi2020estimating} where the original influence problem is reformulated to reduce down to gradient-based similarity.~\citet{sogaard2021revisiting} show that TracIn is more robust and accurate compared to older second-order methods. Beyond the traditional use case of fine-tuning for smaller models, TracIn is successfully applied to augmenting task data from the pretraining data~\cite{han2022orca} and for fact-tracing in large language models~\cite{akyurektowards}.

\paragraph{Dialog safety}
Toxicity detection is a long-standing problem in NLP~\cite{wulczyn2017ex}.
Whereas earlier approaches rely on decision trees and support vector machines~\cite{banik2019toxicity}, state-of-the-art classifiers use deep architectures such as transformers~\cite{caselli2020hatebert, zhou2021challenges}.
With the rise of general purpose chatbots~\cite{openai2022chatgpt}, particular attention has been paid to the problem of toxicity detection in dialog contexts, many featuring adversarial examples deliberately crafted to trick chatbots~\cite{parlai2017, xu2021bot}.

\begin{table*}[t]
\centering
\resizebox{\textwidth}{!}{
    \begin{tabular}{llcc}
    \toprule
    & & \textbf{Label} & \textbf{Prediction} \\ \midrule
    \textbf{Misclassified example (validation)} & \textit{that sounds fun my girlfriend loves to hunt} & OK & Not OK \\
    \textbf{Most influential example (train)} & \textit{that sounds like bad coffee} & Not OK & --- \\ \\
    \textbf{Misclassified example (validation)} & \textit{at 32 years of age i cant say i can do anything french except fries} & Not OK & OK \\
    \textbf{Most influential example (train)} & \textit{i have great memories of sleeping with my cousin in new hampshire} & OK & --- \\\\
    \textbf{Misclassified example (validation)} & \textit{bathing is overrated , i do it once a month} & OK & Not OK \\
    \textbf{Most influential example (train)} & \textit{i am a plumber such a dirty job but somebody gotta do it} & Not OK & --- \\
    \bottomrule
    \end{tabular}
}
\caption{Illustration of misclassified validation set instances and their most influential training set examples as measured using TracIn. The labels \emph{OK} and \emph{Not OK} correspond to not offensive and offensive, respectively.}
\label{tab:parlai_bad_training_examples}
\end{table*}

\section{Influence functions and TracIn}
\label{sec:tracin_background}
Given a training example $z = (x, y)$ and a test example $z_{test} = (x_{test}, y_{test})$, influence functions estimate the change in $L(z_{test})$ (the test example loss) caused by the training example $z$. 
Earlier influence function work~\cite{koh2017understanding} computes this by perturbing the training example around the converged checkpoint and measuring the effect this has on $L(z_{test})$ through changes in the parameters. This essentially comes down to a second order approximation of loss: 
\begin{equation*}
    I(z, z_{test}) = -\nabla_{W}L(z_{test},\hat{W})H_{\hat{W}}^{-1}\nabla_W L (z,\hat{W})
\end{equation*}
where $H$ is the Hessian of the loss at the final model checkpoint.

In this paper, we use an approach from a more recent method that is less computationally expensive and shows promising results~\cite{sogaard2021revisiting, han2022orca}. 
TracIn formulates the question of attribution as an accounting task throughout training. Every time a training example is seen, it records the change in loss for each test example and accumulates the losses throughout training. 
Then, it approximates the loss with a first-order Taylor series:
\begin{equation*}
    I(z, z_{test}) = -\nabla_{W}L(z_{test},\hat{W})^T\nabla_W L (z,\hat{W})
\end{equation*}
The total accounting cost thus reduces down to computing gradients over a set of checkpoints. 
When gradient similarity is used in this form, outlier examples with abnormally large gradients may dominate the retrieval results. 
We use cosine similarity to alleviate this effect, following~\citet{barshan2020relatif}.
We observe that normalized retrieval tends to return examples that are more semantically related.\footnote{We report on additional experiments computing similarities using the dot product without normalization in Section~\ref{sec:similarity_measure}.}

\section{TracIn for soft prompts}
As mentioned, measuring data influence through TracIn is achieved by computing gradient similarities between a training and a test example.
For fine-tuned billion parameter models, this involves computing and saving gradients of the size of the model (number of trainable parameters) per example. While this is intractable without approximation, we utilize parameter-efficient tuning methods, i.e., prompt-tuning~\cite{lester2021ptune}, to reduce the computational cost. 
Since prompt-tuning updates only a small subset of parameters during training (i.e., thousands), our gradient representations are low-dimensional (768 for \tfivebase{}, 4,096 for \tfivexxl{}, and 8,192 for \palm{}) and we can easily measure their similarities. 
It is therefore possible to precisely compute the influence of thousands of training examples on a single test example efficiently by simply measuring vector similarities for the samples' prompt gradients.

To test this method, we train a soft prompt to classify offensive examples in the \textsc{ParlAI} dataset~\cite{dinan-etal-2019-build}, a popular open source dialog toxicity classification dataset comprised of conversational statements and labels indicating whether the statement would be acceptable in friendly conversation. 
We then evaluate our model on the test set, and use TracIn to find the closest, i.e., most influential training set examples for misclassified validation set examples. 
Table~\ref{tab:parlai_bad_training_examples} shows a sample of such pairs of misclassified validation set examples and their most influential training set examples. 
According to these results the misclassifications may not indicate a failure by the model to learn the task, but rather the existence of questionably labeled examples in the training set.

\subsection{\gbair{}}
Having established the advantages of measuring data influence efficiently using prompt-tuning, we here explain how this approach can be used to identify and mitigate corrupt training examples. 

We propose \textit{Gradient-Based Automated Iterative Recovery} (\gbair{}) for parameter-efficient tuning---a protocol for identifying and relabeling mislabeled training examples in a dataset. 
\gbair{} is meant to be applied iteratively to a training set over a number of $n$ iterations.

\begin{algorithm}[!ht]
\caption{\textit{Gradient-Based Automated Iterative Recovery} (\gbair{})}\label{alg:cap}
\begin{algorithmic}
\Require Language model $L$, training set $T_{train}$, validation set $T_{val}$, number of iterations $n$, number of influential examples to consider $k$, number of examples to relabel $\tau$
\State $T^{1}_{train} \leftarrow T_{train}$
\For{$i \in \{1, ..., n\}$}
\State $p \leftarrow \texttt{train\_prompt}(L, T^{i}_{train})$
\State $V^{i} \leftarrow \texttt{sample\_validation\_set}(T_{val})$
\State $V^{i}_{mis} \leftarrow \texttt{get\_misclassified}(L, p, V^{i})$ 
\State $T^{i}_{inf} \leftarrow \texttt{get\_inf}(L, p, T^{i}_{train}, V^{i}_{mis}, k, \tau)$
\State $T^{i}_R \leftarrow \texttt{relabel\_examples}(T^{i}_{inf})$
\State $T^{i+1}_{train} \leftarrow T^{i}_{train} \setminus T^{i}_{inf} \cup T^{i}_{R}$
\EndFor
\end{algorithmic}
\label{alg:g-bair}
\end{algorithm}

The method is illustrated in Algorithm~\ref{alg:g-bair}. Suppose we are given a language model $L$, a training set $T_{train}$ containing a fraction of mislabeled examples, as well as a validation set $T_{val}$ and a test set $T_{test}$ containing only correctly labeled examples. 
In each iteration $i$, \gbair{} uses TracIn to identify influential training set examples for misclassified validation set examples. 
To do so, we first train a prompt $p$ on the training set $T^{i}_{train}$ using language model $L$ (\texttt{train\_prompt}). 
We then sample a validation subset $V^i$ from $T_{val}$ (\texttt{sample\_validation\_set}) and run inference over it, retaining only the misclassified instances from the validation set, denoted $V^{i}_{mis}$ (\texttt{get\_misclassified}). 
Using TracIn, we compute the $k$ most influential training set examples for each example in $V^{i}_{mis}$, and rank the retrieved influential examples according to their frequency (\texttt{get\_inf}). 
We then consider the set $T^{i}_{inf}$ containing the $\tau$ most commonly occurring influential examples to be mislabeled, and relabel them to obtain $T_R^{i}$ (\texttt{relabel\_examples}).\footnote{Since we only consider binary datasets in our experiments, relabeling is achieved by swapping the label.}
This set is used to modify the training set by removing $T^{i}_{inf}$ and adding $T^{i}_{R}$. Afterwards, we retrain the prompt $p$ on the modified training set $T_{train}^{i+1}$. 

Following this protocol over multiple iterations, we assess model performance using the prompt at each iteration on the held-out test set $T_{test}$.

\section{Experiments}
To assess \gbair{}'s performance at identifying and mitigating mislabeled training data, we report on a series of experiments using manually corrupted datasets.

\subsection{Models}
We conduct our experiments on three pretrained language models and further prompt-tune them with the datasets described in Section~\ref{sec:datasets}. The first two are variants of T5~\cite{raffel2020exploring}, namely the \textsc{Base} version with 220 million parameters and the XXL version with 11 billion parameters. 
The third is the 62 billion parameter version of \textsc{PaLM}~\cite{chowdhery2022palm}. 
We decided to use these three models in order to test whether there may exist a correlation between model size and TracIn performance. 

Across experiments, we tune soft prompts consisting of 10 token vectors, using the Adam optimizer~\cite{kingma2014adam} with a learning rate of 0.1 and a weight decay of 0.0001.
For T5 models we use a batch size of 32, and for \palm{} one of 4.\footnote{The \palm{} model is very large so memory constraints limit the batch size we were able to use during prompt-tuning.} 
We train all models for 20 epochs.

\subsection{Datasets}
\label{sec:datasets}
We experiment with two datasets from the  \textsc{ParlAI}~\cite{dinan-etal-2019-build} data collection effort, denoted \parlaistandard{} and \parlaiadversarial{}.
The \textsc{ParlAI} datasets consist of single-turn conversations annotated based on offensiveness. 
For the \textsc{Standard} dataset, crowdworkers were simply asked to write sentences that they would consider offensive, whereas for the \textsc{Adversarial} one, the workers were asked to write sentences that are offensive, but that a classifier might predict to be safe. 
Both datasets come with pre-defined splits of 24,000 examples for training, and 3,000 each for validation and testing. 

All three language models perform well on the test set portions of the two datasets when prompt-tuned with random samples of 1,000 examples from the training sets. During sampling, we ensure that the resulting training set is class-balanced (the validation and test sets are imbalanced with positive examples making up around 10\% of the data).

\begin{figure}[!ht]
     \centering
     \begin{subfigure}[b]{0.98\columnwidth}
         \centering
         \includegraphics[width=\columnwidth]{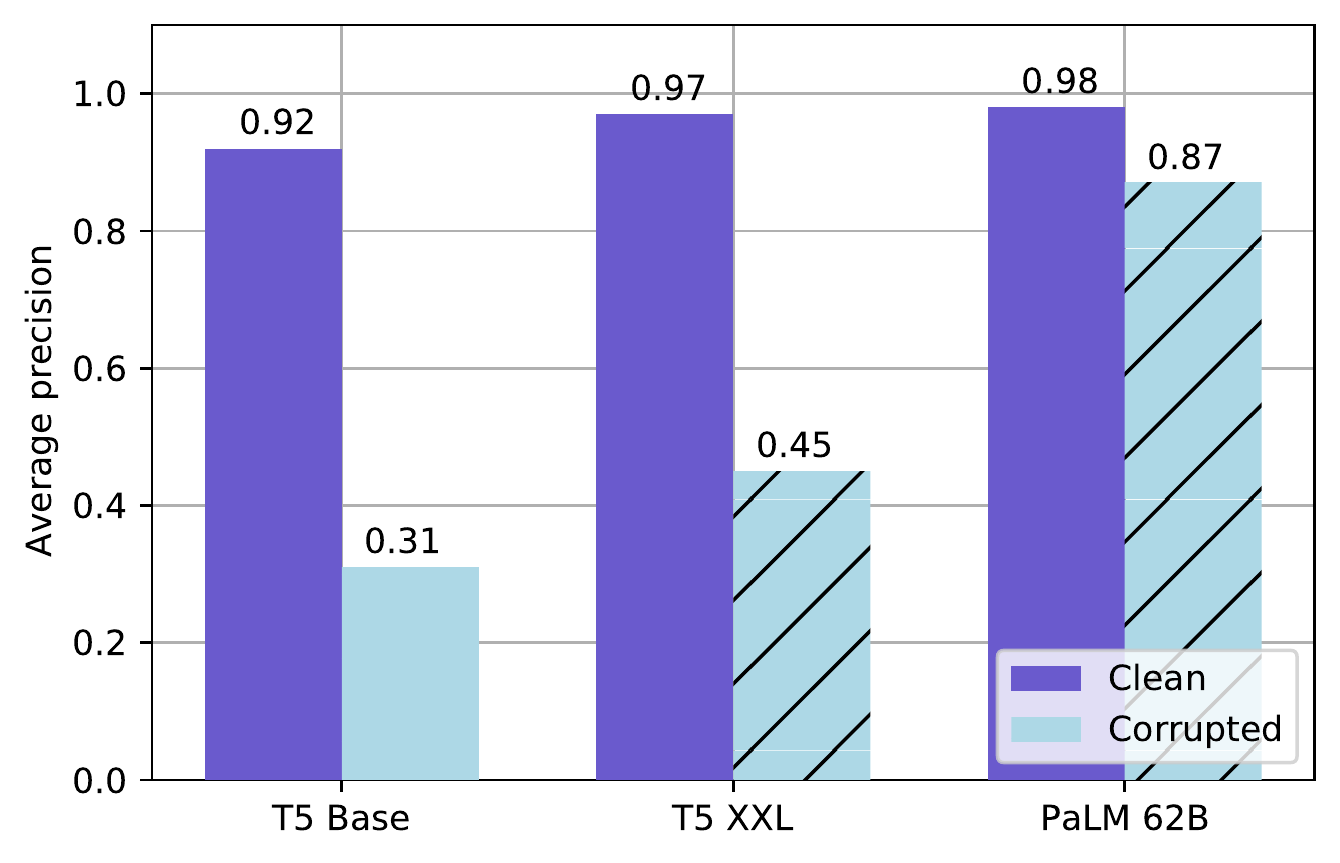}
         \caption{\parlaistandard{}}
     \end{subfigure}
     \hfill
     \begin{subfigure}[b]{0.98\columnwidth}
         \centering
         \includegraphics[width=\columnwidth]{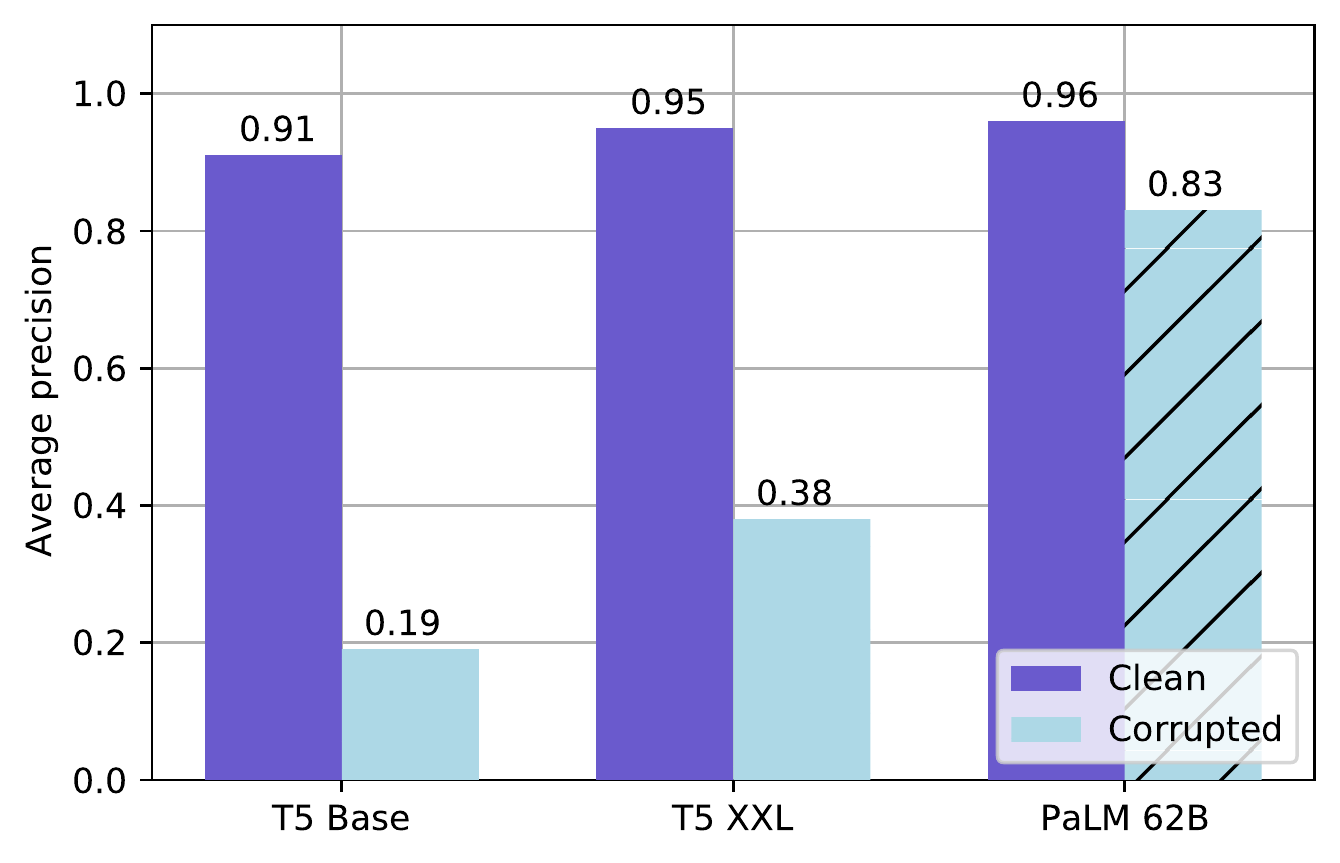}
         \caption{\parlaiadversarial{}}
     \end{subfigure}
     \hfill
    \caption{Illustration of prompt-tuning model performances (in terms of average precision) before (Clean) and after corrupting 30\% of the training data (Corrupted) for both \parlaistandard{} (a) and \parlaiadversarial{} (b).}
    \label{fig:standard_corrupted_performance}
\end{figure}

\subsection{Experimental details}
To evaluate our method, we manually corrupt the dataset by randomly flipping 30\% of the labels (we denote the corrupted training set as $T_{train}^C$). 
We found this level to be sufficient for causing a significant drop in model accuracy and compare the effects of choosing different levels of corruption in ablation studies (Section~\ref{subsec:ablation_corruption_rate}). 
Figure~\ref{fig:standard_corrupted_performance} shows the impact on performance for each model as a result of this corruption. 
Then we train a classifier on the corrupted dataset, and evaluate on the validation data. 
Using \gbair{} for $n=10$ iterations, we take misclassified examples in the validation set, and identify their most influential training set examples according to TracIn. 
We collect the $\tau=20$ most frequently identified training set examples according to this method (aggregated from the $k=3$ most influential training set examples for each misclassified validation set example) in each iteration, and relabel them. 
We furthermore select a subset of 500 examples from the entire validation set (containing 3,000 examples each for \parlaistandard{} and \parlaiadversarial{}) at each iteration to form $V^i$.

Our method aims to iteratively clean up the dataset by repeating this intervention, retraining the classifier each time. 
For the prompts trained in each iteration, we sample 200 examples from the datasets' validation sets for checkpoint evaluation. 
Specifically, after each epoch during prompt-tuning, we evaluate performance on the sampled validation set, and select the checkpoint producing the lowest loss for testing.

\paragraph{Baselines} We compare this intervention to two baselines: (1) randomly removing $\tau=20$ training set examples, and (2) removing the $\tau=20$ training set examples that are semantically closest to misclassified validation set examples in embedding space, computed with \sentencetfive{}~\cite{ni2021sentence}.\footnote{We conducted additional experiments using the universal sentence encoder~\cite{cer2018universal}, but did not notice any substantial performance differences. Results for these experiments can be found in Appendix~\ref{app:baseline_use}.} 
The embedding similarity baseline lets us study the effect of semantic similarity in isolation, in order to rule out the possibility that model performance on a validation set example can be predicted through its tokens alone. 
\paragraph{Evaluation metrics} We analyze \gbair{} task performance in two ways. 
First, we are interested in the recovery performance, measured in terms of \textbf{\textit{average precision}} (AP), achieved by the method over a range of iterations. 
The AP is computed as the area under the precision-recall curve. 
Second, we analyze the method's \textbf{\textit{\underline{c}orrupted \underline{i}nstances \underline{i}dentification \underline{r}ate}} (CI$^2$R). CI$^2$R measures how many relevant (i.e., via the corruption process mislabeled) training set examples \gbair{} identified over the iterations. 
To do so, we consider the set of corrupted training examples $T_{train}^C$ and measure the fraction of corrupted examples retrieved in $T_{inf}^{i}$ at each iteration $i$, and compute the average over $n$ iterations. 
Formally, 
$$
\text{CI$^2$R} = \frac{1}{n} \sum_{i=1}^n \frac{|T_{inf}^{i}\cap T_{train}^C|}{|T_{inf}^{i}|}
$$
The CI$^2$R lies between a value of 0.0 and 1.0, with the former representing a total miss (i.e., none of the identified influential examples have been corrupted) and the latter representing a total hit (i.e., all of the identified influential examples have been corrupted).

\subsection{Results}

\begin{table*}[!ht]
    \centering
    \resizebox{\textwidth}{!}{
    \begin{tabular}{c l c c c c c c c c}
    \toprule
    & & \multicolumn{5}{c}{\textbf{AP}} & \multicolumn{3}{c}{\textbf{CI}$^2$\textbf{R}} \\ 
    \cmidrule(lr){3-7} \cmidrule(lr){8-10}
    \textbf{Dataset} & \textbf{Model} & \textbf{Clean} & \textbf{Corrupted} & \random{} & \sentencetfive{} & \gbair{} & \random{} & \sentencetfive{} & \gbair{} \\
    \midrule
    \multirow{3}{*}{\parlaistandard{}} 
    & \tfivebase{} & $0.92_{0.01}$ & $0.31_{0.09}$ & $0.39_{0.05}$ & $0.43_{0.07}$ & $\boldsymbol{0.61}_{0.08}$ & $0.20_{0.01}$ & $0.18_{0.03}$ & $\boldsymbol{0.43}_{0.01}$ \\
    & \tfivexxl{} & $0.97_{0.00}$ & $0.45_{0.12}$ & $0.36_{0.03}$ & $0.39_{0.07}$ & $\boldsymbol{0.76}_{0.11}$ & $0.21_{0.03}$ & $0.22_{0.06}$ & $\boldsymbol{0.53}_{0.02}$ \\
    & \palm{} & $0.98_{0.00}$ & $0.87_{0.08}$ & $0.90_{0.04}$ & $0.92_{0.02}$ & $\boldsymbol{0.93}_{0.02}$ & $0.20_{0.00}$ & $0.22_{0.02}$ & $\boldsymbol{0.40}_{0.01}$ \\
    \midrule
    \multirow{3}{*}{\parlaiadversarial{}} 
    & \tfivebase{} & $0.91_{0.03}$ & $0.19_{0.03}$ & $0.27_{0.08}$ & $0.29_{0.07}$ & $\boldsymbol{0.54}_{0.05}$ & $0.21_{0.03}$ & $0.19_{0.06}$ & $\boldsymbol{0.39}_{0.02}$ \\
    & \tfivexxl{} & $0.95_{0.01}$ & $0.38_{0.15}$ & $0.36_{0.09}$ & $0.49_{0.02}$ & $\boldsymbol{0.73}_{0.13}$ & $0.21_{0.01}$ & $0.22_{0.03}$ & $\boldsymbol{0.38}_{0.14}$ \\
    & \palm{} & $0.96_{0.00}$ & $0.83_{0.04}$ & $0.83_{0.04}$ & $0.82_{0.01}$ & $\boldsymbol{0.90}_{0.01}$ & $0.20_{0.01}$ & $0.21_{0.03}$ & $\boldsymbol{0.39}_{0.04}$ \\
    \bottomrule
    \end{tabular}
    }
    \caption{Mean (standard deviation) performance scores in terms of average precision (AP) as well as the CI$^2$R for clean, corrupted, and recovered training sets across three seeds. For AP, \textbf{Clean} and \textbf{Corrupted} denote performances on the test set before and after corrupting 30\% of the training data. \random{} and \sentencetfive{} show the recovered performances using the two baselines, and \gbair{} shows recovered performance using our proposed method. Best performances per metric and model-dataset combination are highlighted in bold.}
    \label{table:main_tracin_results}
\end{table*}

\paragraph{Average precision} Performance results using the two baselines (\random{} and \sentencetfive{}) as well as \gbair{} in terms of average precision can be found in Table~\ref{table:main_tracin_results} (area denoted with AP). We report the clean model performance (\textbf{Clean}), the performance after corruption (\textbf{Corrupted}), and the best performance achieved after iterating over and relabeling examples in the dataset using the three methods. We first observe that both \tfivebase{} and \tfivexxl{} exhibit substantial hits in performance after corruption (e.g., from $0.91$ AP to $0.19$ AP for \tfivebase{} and $0.95$ AP to $0.38$ AP for \tfivexxl{} on \parlaiadversarial{}). This is in contrast to \palm{}, which shows less substantial decreases in performance after corruption. 

For both \tfivebase{} and \tfivexxl{}, we observe that across both datasets, \gbair{} largely outperforms the two baselines in terms of AP recovery over iterations, recovering up to 35\% AP (\tfivebase{} on \parlaiadversarial{} with $0.19\,\text{AP} \to 0.54\,\text{AP}$ and \tfivexxl{} on the same dataset with $0.38\,\text{AP} \to 0.73\,\text{AP}$). The \sentencetfive{} baseline seems to provide little additional benefits over the \random{} baseline (with the exception of \tfivexxl{} on \parlaiadversarial{}, where we observe a difference of $0.13$ AP between the two baselines), indicating that relabeling instances based on semantic similarity cannot recover the performance drop incurred through training data corruption.

The performance recovery for \palm{} is less clear as compared to T5. 
We do observe that for both datasets, \gbair{} outperforms baselines in terms of AP recovery, yet the differences between baselines and \gbair{} is at only $0.01$ AP in absolute value for \parlaistandard{}, and $0.07$ AP for \parlaiadversarial{}. Given the relatively small drop in performance after corruption ($0.98$ AP $\to$ $0.87$ AP for \parlaistandard{} and $0.96$ AP $\to$ $0.83$ AP for \parlaiadversarial{}), these results might not be unexpected. The larger model seems less affected by mislabeled examples, a result also observed of in-context learning~\cite{min2022rethinking}: it performs well even after a 30\% corruption, thus mislabeled training examples seem to play a less impactful role for model decision making, and mitigating their existence is hence less impactful to the resulting model AP performance scores.

\begin{figure*}[!ht]
     \centering
     \begin{subfigure}[b]{0.48\textwidth}
         \centering
         \includegraphics[width=\textwidth]{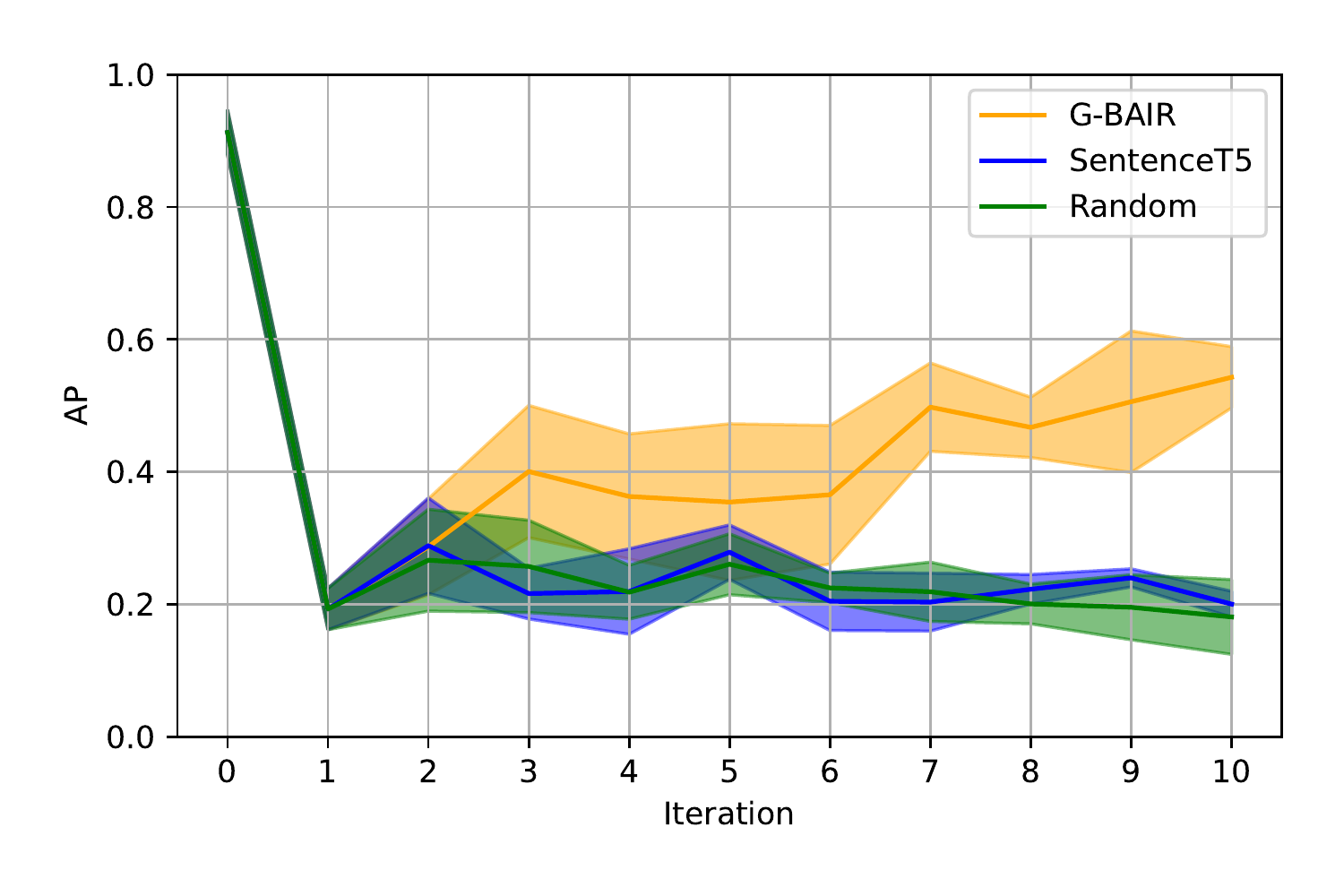}
         \caption{TracIn performance recovery}
     \end{subfigure}
     \hfill
     \begin{subfigure}[b]{0.48\textwidth}
         \centering
         \includegraphics[width=\textwidth]{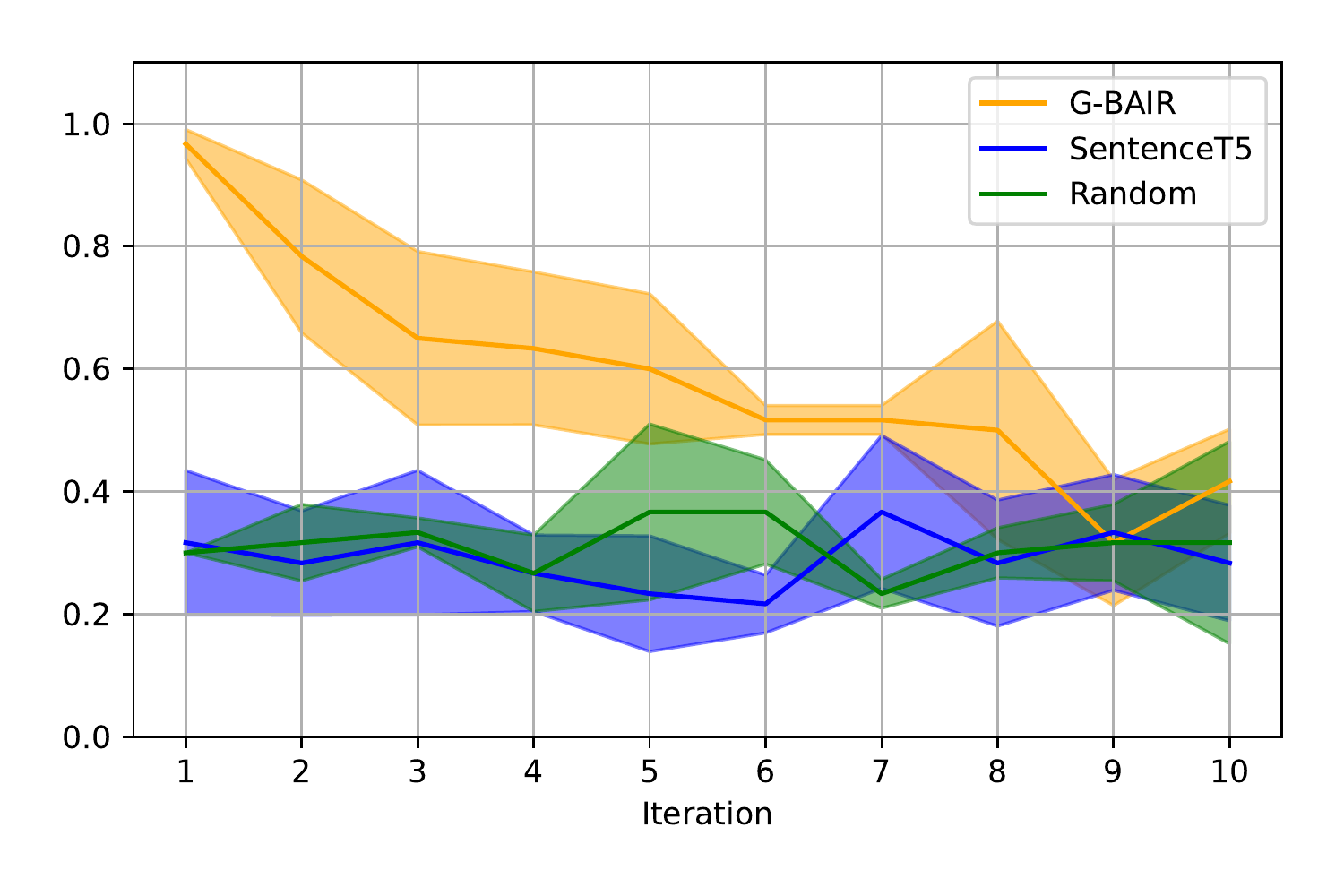}
         \caption{Fraction of corrupted examples identified by TracIn}
     \end{subfigure}
     \hfill
    \caption{Illustration of model performance recovery for \tfivebase{} on \parlaiadversarial{} in terms of AP (a) and the fraction of identified corrupted examples per iteration (b). Results are averaged across three independent runs with the standard deviations shown.}    
    \label{fig:plot_tfive_base_adversarial}
\end{figure*}

\paragraph{CI$^2$R.}Performance results in terms of CI$^2$R can be found in Table~\ref{table:main_tracin_results} (area denoted CI$^2$R). Here we report the CI$^2$R for both baselines as well as \gbair{} across models and datasets. In line with the AP recovery results, we observe that \gbair{} largely outperforms both baselines in terms of CI$^2$R. We observe that both the \random{} and \sentencetfive{} baselines exhibit scores of around 0.2 consistently across experiments. For the former this is expected: the \random{} baseline relabels 20 training examples in each iteration, of which 6 (i.e., 30\% of 20) are on average misclassified. After $n=10$ iterations the baseline has then relabeled $10 \cdot 6 = 60$ mislabeled examples, which makes up 20\% of the 300 corrupted training examples. It is interesting to see that the \sentencetfive{} baseline does not provide any additional benefit in terms of CI$^2$R over the \random{} one. \gbair{}, however, exhibits CI$^2$R scores far above the random draw, with scores reaching up to 0.52 (\tfivexxl{} and \parlaistandard{}). This demonstrates that \gbair{} is able to use TracIn effectively to identify corrupted training examples, and gradients encode extra information that is not present in embeddings.

Figure~\ref{fig:plot_tfive_base_adversarial} illustrates the recovery performance for \gbair{} and the two baselines with respect to the AP (a) and the fraction of identified corrupted training examples per iteration (b). For (a), iteration 0 denotes model test set performance when trained on the clean training set and iteration 1 when trained on the corrupted training set. Iterations 2--10 then show the performance recovery for each method. 
As we can see, \gbair{} shows clear improvements with respect to both evaluation settings. For (b), we additionally observe that in the first iteration, close to 100\% of the influential examples identified by \gbair{} were indeed corrupted. The fraction of identified corrupted examples gradually decreases with an increasing number of iterations, indicating that an increasing test set performance yields a decrease in the retrieval of corrupted influential examples. Additional figures illustrating the remaining experiments can be found in Appendix~\ref{app:tracin_plots}.

\section{Ablation studies}
\label{sec:ablation_studies}
We conduct a series of additional analyses to better understand the impact of validation set size, corruption rate, similarity measure, and intervention method for influential examples when using \gbair{}.

\begin{figure}[!ht]
    \centering
    \includegraphics[width=\columnwidth]{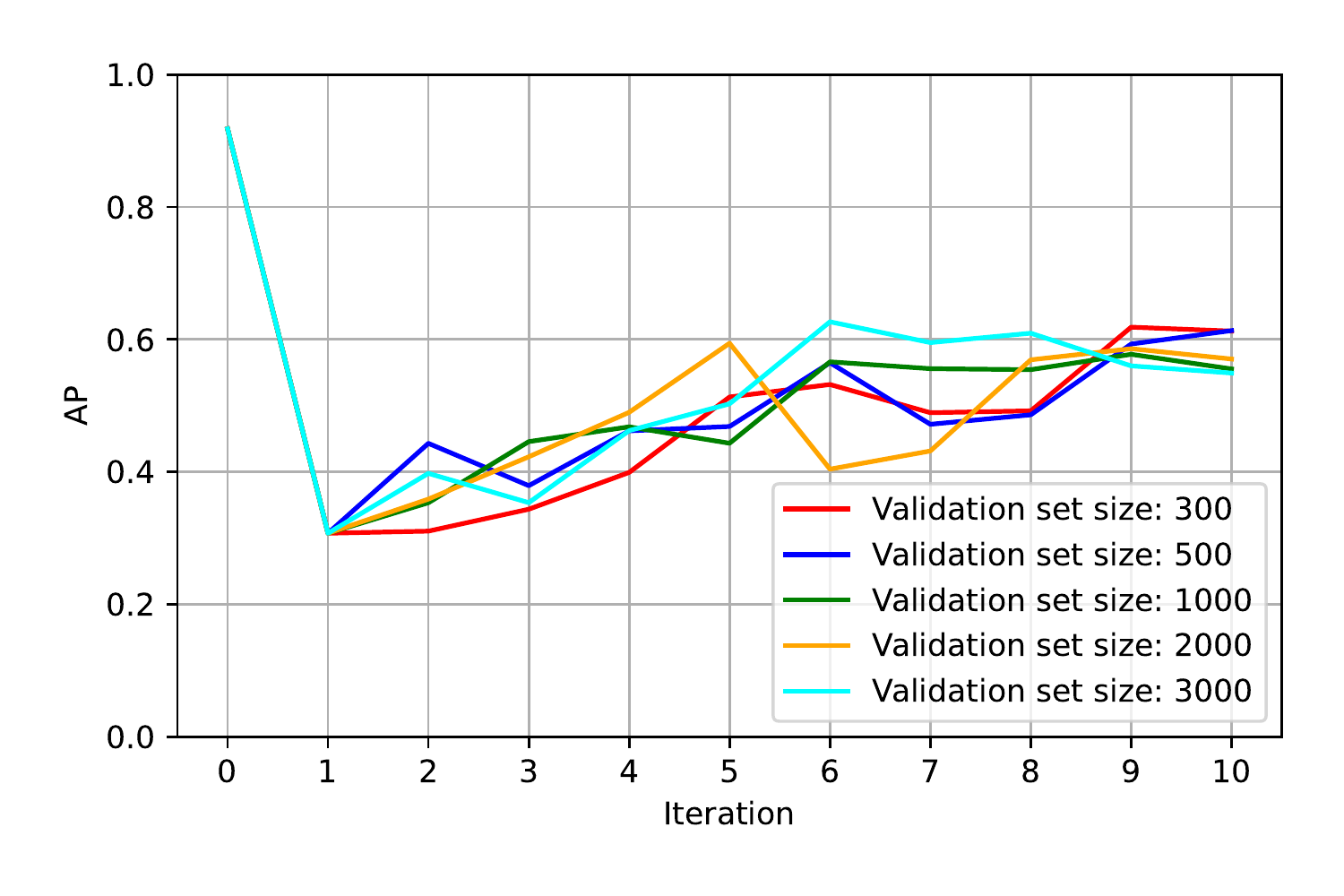}
    \caption{Comparison of \gbair{} recovery performance for different validation set sizes (300, 500, 1,000, 2,000, 3,000) when run on \parlaistandard{} with \tfivebase{}. We show average results across three seeds.}
    \label{fig:ablation_val_set_size}
\end{figure}

\subsection{Different validation set sizes}
\label{subsec:ablation_validation_size}
We first investigate the impact of the validation set size on the recovery rate of \gbair{}. To do this, we experiment with validation set sizes of 300, 1,000, 2,000, and 3,000 in addition to the 500 as shown above. Experiments are conducted with \tfivebase{}, on \parlaistandard{}. 

Results can be seen in Figure~\ref{fig:ablation_val_set_size}.  We observe that performance recovery does not seem to dramatically differ between different validation set sizes. This is somewhat unexpected, since one could argue that a larger validation set size leads to a larger absolute number of misclassified validation set instances (for a fixed model performance), which in turn creates a larger pool of influential training examples that may better represent the corrupt training set. However, the experimental results hint at a different picture. It seems that even with a validation set of 300 examples, \gbair{} is capable of identifying a reasonable set of corrupted examples, which, when removed from the training set, leads to notable performance recovery on the test set. This finding suggests that \gbair{} may be useful even without a large validation set.\footnote{For many datasets, including \textsc{ParlAI}, a proportion of the validation data can also be noisy. Our results use the validation data as is and therefore serve as a lower bound to the performance. \gbair{} also helps in settings where validation data can be very noisy, since cleaning a small set of examples is easier than cleaning the whole training data.} 

\begin{figure}[!ht]
    \centering
    \includegraphics[width=\columnwidth]{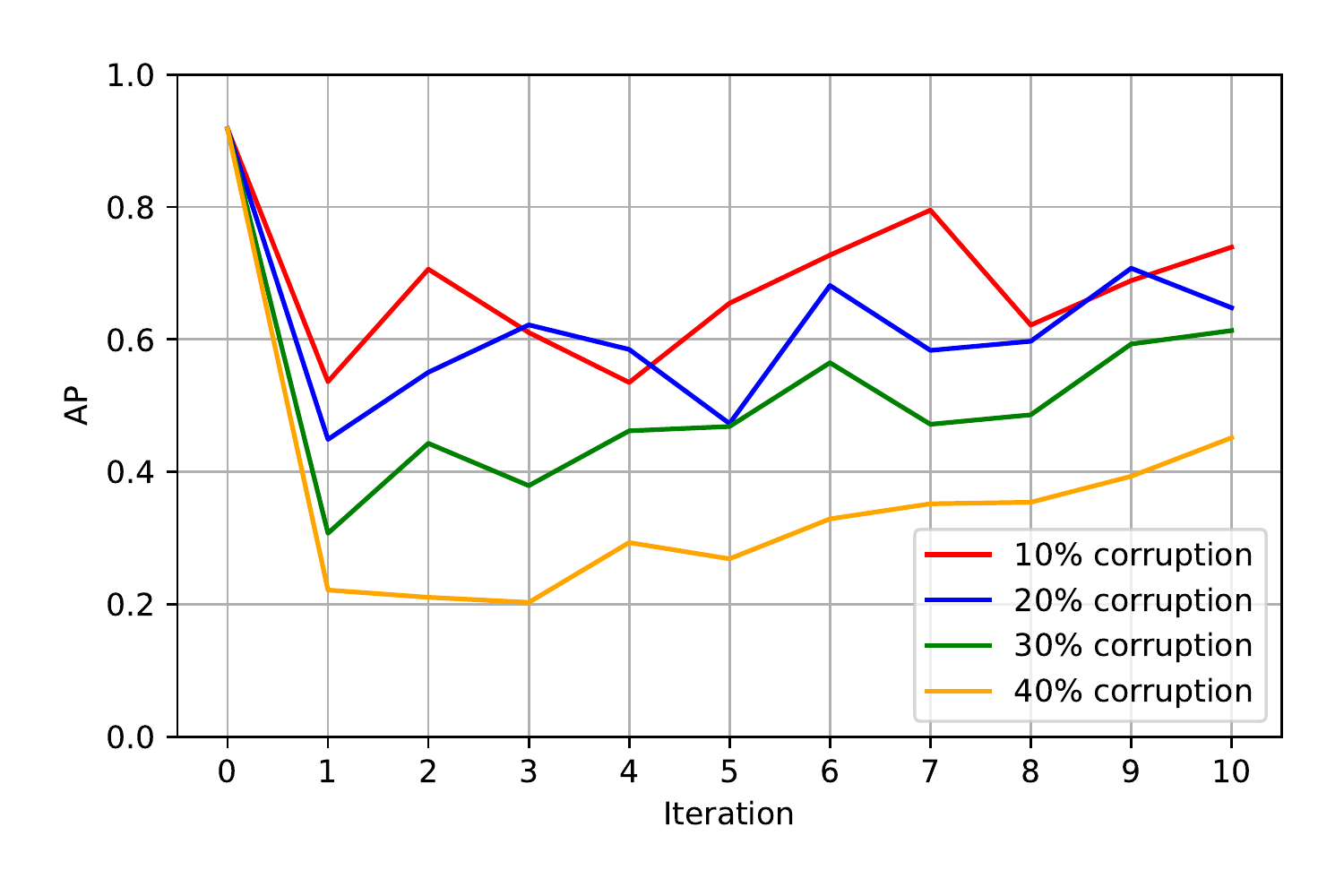}
    \caption{Comparison of \gbair{} recovery performance for different corruption rates (10\%, 20\%, 30\%, 40\%) when run on \parlaistandard{} with \tfivebase{}. We show average results across three seeds.}
    \label{fig:ablation_corruption_rate}
\end{figure}

\subsection{Corruption rate}
\label{subsec:ablation_corruption_rate}
We furthermore experiment with \tfivebase{} on \parlaistandard{} using different corruption rates, i.e., 10\%, 20\%, and 40\% in addition to the results with 30\% shown above. 

The results can be found in Figure~\ref{fig:ablation_corruption_rate}. It can be seen that the larger the corruption rate, the larger the initial drop in performance on the test sets. However, across corruption rates, we observe that \gbair{} is able to successfully recover performances, indicating that the method is able to identify mislabeled data and mitigate their harms even in the presence of a smaller number (i.e., 10\%) of corrupted examples.

\begin{figure}[!ht]
    \centering
    \includegraphics[width=\columnwidth]{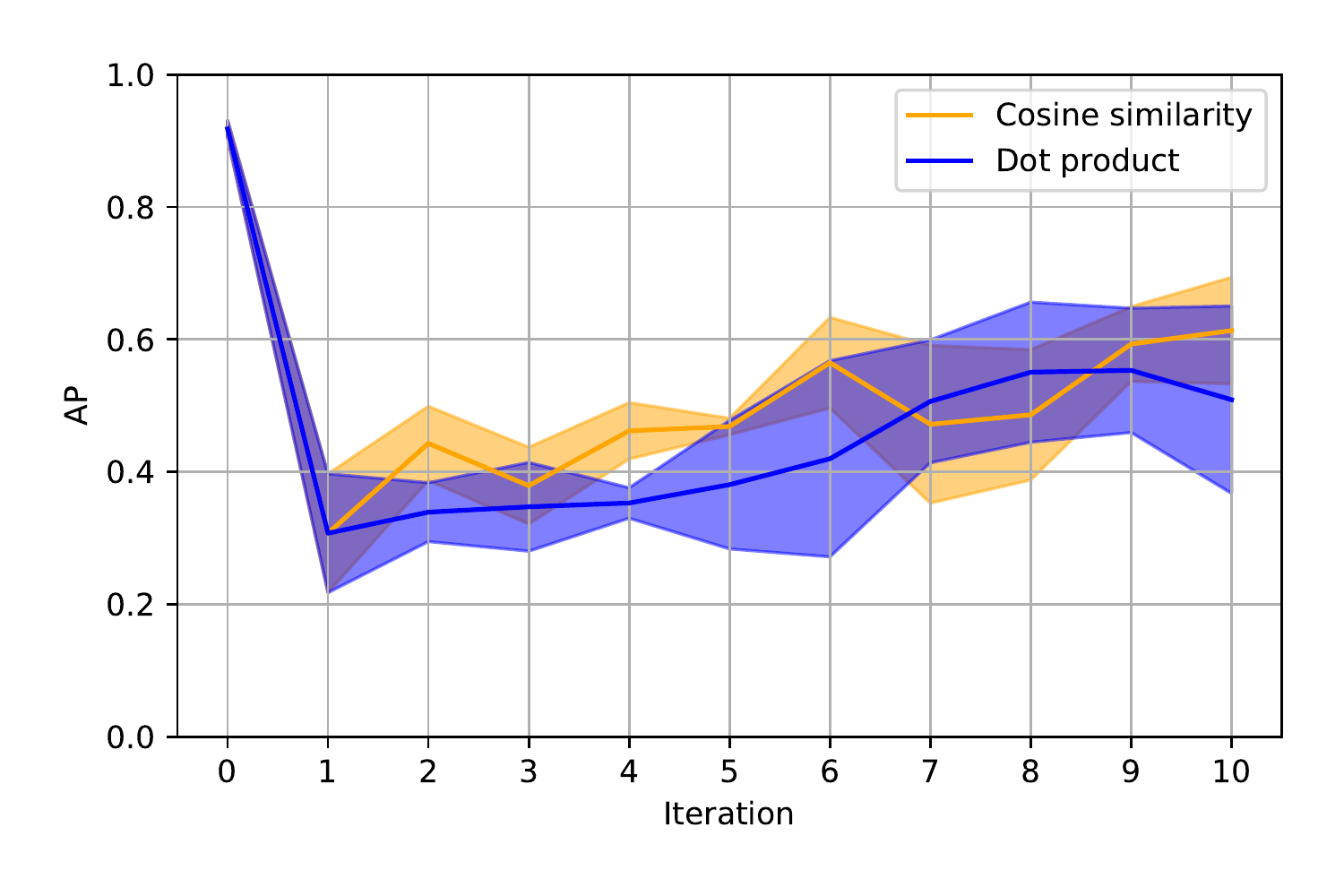}
    \caption{Comparison of \gbair{} recovery performance with two different similarity measures (cosine similarity and dot product) when run on \parlaistandard{} with \tfivebase{}. We show average results across three seeds with their respective standard deviations.}
    \label{fig:ablation_similarity_measure}
\end{figure}

\subsection{Similarity measure}
\label{sec:similarity_measure}
We also study whether using the dot product, i.e., the unnormalized cosine distance, as an alternative measure of similarity might have an impact on the recovery performance. As mentioned in Section~\ref{sec:tracin_background}, using unnormalized measures of similarity between gradient representations may lead to the retrieval of outlier examples with large gradient magnitudes, and could potentially hinder the effects obtained from relabeling influential examples. In line with previous experiments, we report results using \tfivebase{} on \parlaistandard{}.

The results in Figure~\ref{fig:ablation_similarity_measure} show that in practice, the choice of similarity measure seems to make little difference with respect to \gbair{} recovery performance. We observe that both measures yield similar recovery results. The standard deviations obtained using the dot product tend to be slightly larger as compared to the cosine similarity. This could be explained through the aforementioned argument that unnormalized measures of similarity might retrieve a smaller, more concentrated set of influential examples with large gradient magnitudes. This might result in worse generalization after the relabeling process. 

\begin{figure}[!ht]
    \centering
    \includegraphics[width=\columnwidth]{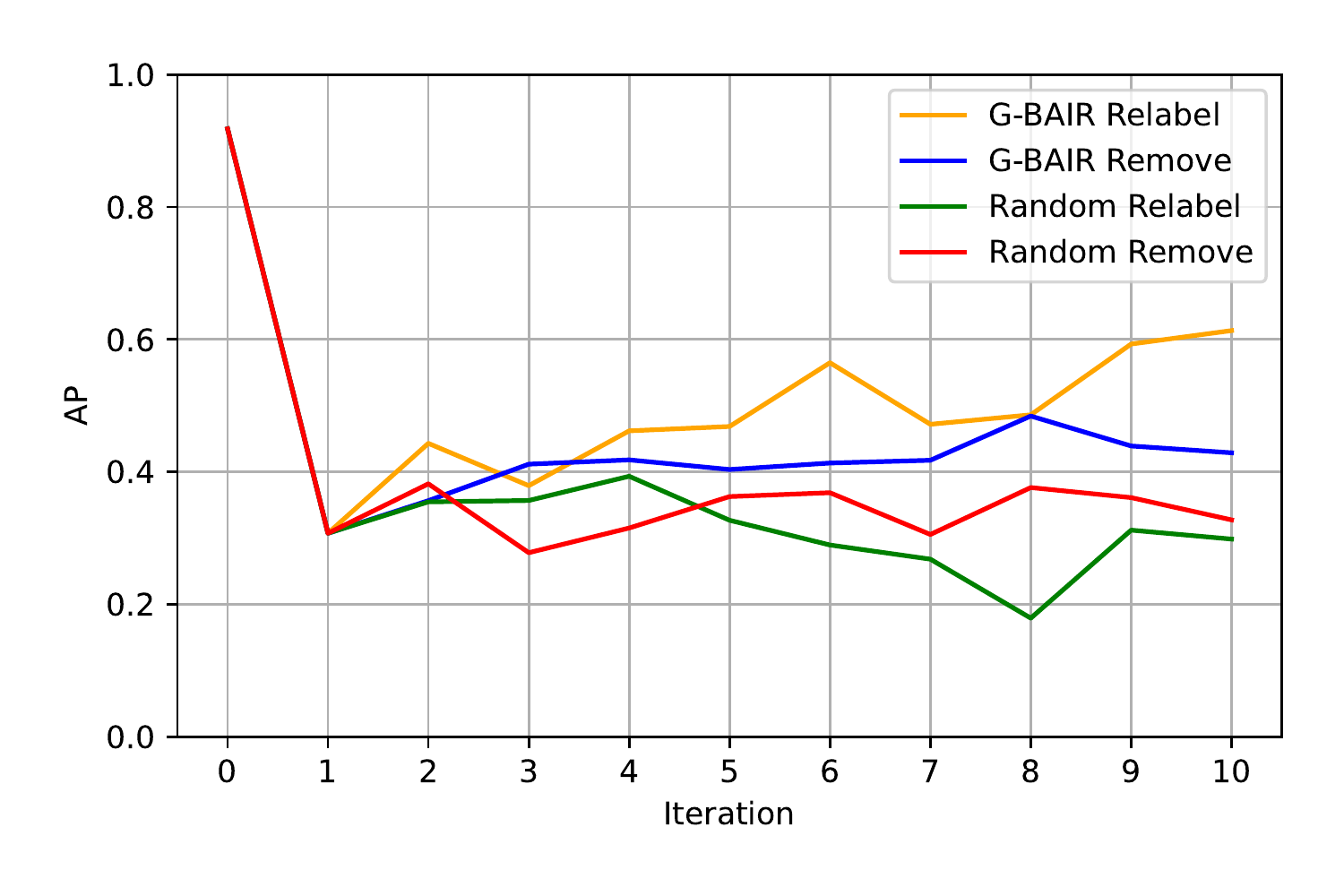}
    \caption{Comparison of \gbair{} recovery performance with two different recovery intervention methods (relabeling and removing) when run on \parlaistandard{} with \tfivebase{}. We show average results across three seeds.}
    \label{fig:ablation_relabeling_removing}
\end{figure}

\subsection{Relabeling or removing instances}
Finally, we repeat an experiment with \tfivebase{} on \parlaistandard{} in which instead of relabeling influential training examples, we remove them from the datasets. Removing examples instead of relabeling them has the advantage that it generalizes to non-binary tasks where easy automated relabeling is not possible. Unlike relabeling, however, removal shrinks the model's training set and might lead to scenarios in which too few training examples remain to fit a model via prompt-tuning. Figure~\ref{fig:ablation_relabeling_removing} shows that, although relabeling tends to work better for \gbair{}, removal performs reasonably well, and we do not observe significant drops in model performance due to smaller training data. The \textbf{Random Remove} baseline yields fairly constant AP scores across iterations, even though 200 training examples (i.e., 20\% of the training set) will have been removed after 10 iterations.

\section{Discussion}
In this paper we introduced \gbair{}, a protocol for iteratively improving the performance of an LLM trained under the PET regime. We showed that gradient-based measures of influence on misclassified validation set examples can identify corruptions in the training set. Finally, we presented effective mitigation strategies that enable LLMs to recover from such corruption at a range of different rates (from 10\% corruption to 40\% corruption). We observed that the model size, and accordingly an increased test set performance on clean data, seems to play a role in the effectiveness of recovery. \palm{}, shown to be robust against a corruption rate of 30\% on the training data (Figure~\ref{fig:standard_corrupted_performance}), exhibited a less clear recovery of AP performance through \gbair{}. Nevertheless, considering performance in terms of CI$^2$R, it is clear that TracIn-based retrieval of influential examples yields far more corrupted examples compared to embedding similarity-based and random baselines.
We also discovered that the model performance can be consistently recovered through \gbair{} across validation set sizes (Section~\ref{subsec:ablation_validation_size}), showing that a few hundred, rather than thousands of validation examples suffice to identify and mitigate corrupted examples in training sets.

A core limiting assumption for our method is that one has access to a golden, correctly labeled validation set. 
This is of course not always the case, but more fundamentally we presume that golden labels are obtainable for one’s task. As LLMs are tasked with increasingly difficult problems, especially ones requiring judgment, the notion of ground truth starts to become elusive~\cite{gordon2021disagreement}. We observed when inspecting training examples from our test domain of conversational safety, that reasonable individuals may have genuine disagreements over the acceptability of an utterance.

We believe a fruitful area of future work is bringing humans into the iteration loop to see whether more sophisticated interventions, beyond simply removing or relabeling examples, could further improve performance.
Flipped labels are only one (straightforward) example of a data quality issue, which lends itself to automated mitigations. 
In the case of a legitimately ambiguous example, human intervention may be the only recourse. 
For example, TracIn may identify confusing examples that could be manually edited to provide more signal to the classifier.
We envision methods like \gbair{} as tools to ultimately empower humans to more quickly diagnose data quality issues.
As methods like parameter-efficient tuning enable us to move toward faster training loops using smaller datasets, data quality becomes even more important, and so do methods for dataset iteration.

\bibliography{anthology,custom}
\bibliographystyle{acl_natbib}

\appendix

\section{Illustrations of performance recovery}
\label{app:tracin_plots}
\begin{figure*}[!ht]
     \centering
     \begin{subfigure}[b]{0.48\textwidth}
         \centering
         \includegraphics[width=\textwidth]{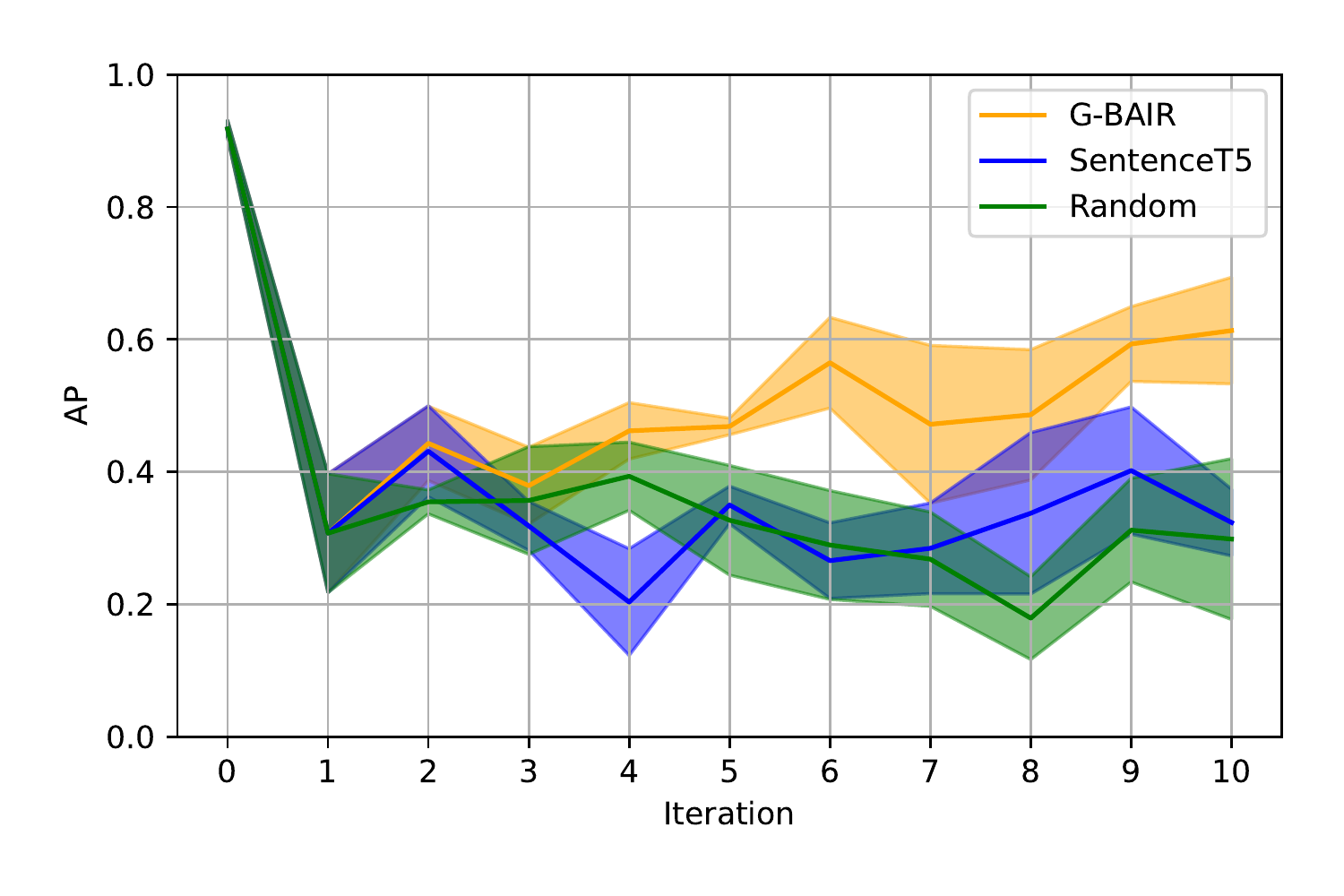}
         \caption{TracIn performance recovery}
     \end{subfigure}
     \hfill
     \begin{subfigure}[b]{0.48\textwidth}
         \centering
         \includegraphics[width=\textwidth]{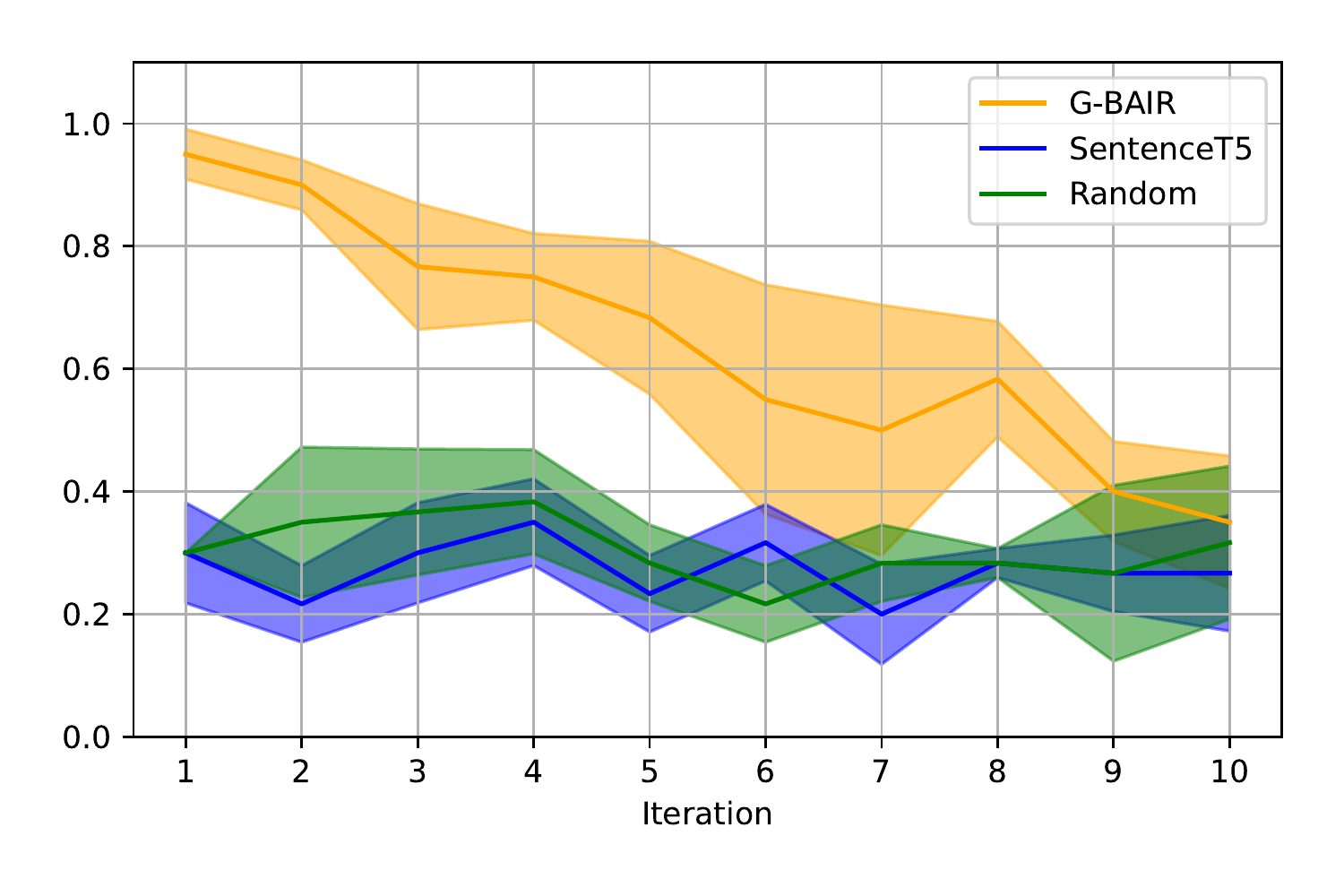}
         \caption{Fraction of corrupted examples identified by TracIn}
     \end{subfigure}
     \hfill
    \caption{Illustration of model performance recovery for \tfivebase{} on \parlaistandard{} in terms of AP (a) and the fraction of identified corrupted examples per iteration (b). Results are averaged across three independent runs with the standard deviations shown.}
    \label{fig:plot_tfive_base_standard}
\end{figure*}

\begin{figure*}[!ht]
     \centering
     \begin{subfigure}[b]{0.48\textwidth}
         \centering
         \includegraphics[width=\textwidth]{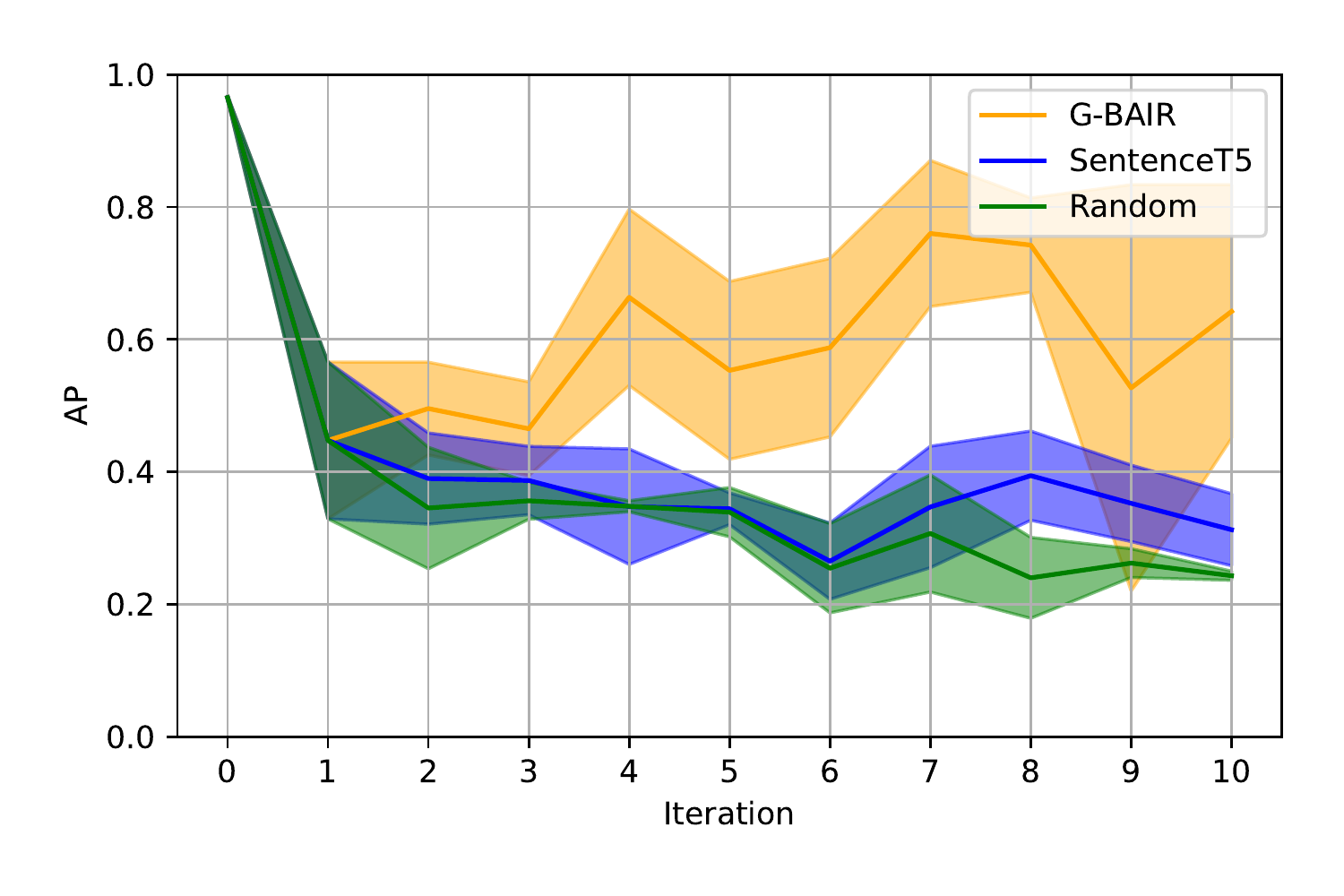}
         \caption{TracIn performance recovery}
     \end{subfigure}
     \hfill
     \begin{subfigure}[b]{0.48\textwidth}
         \centering
         \includegraphics[width=\textwidth]{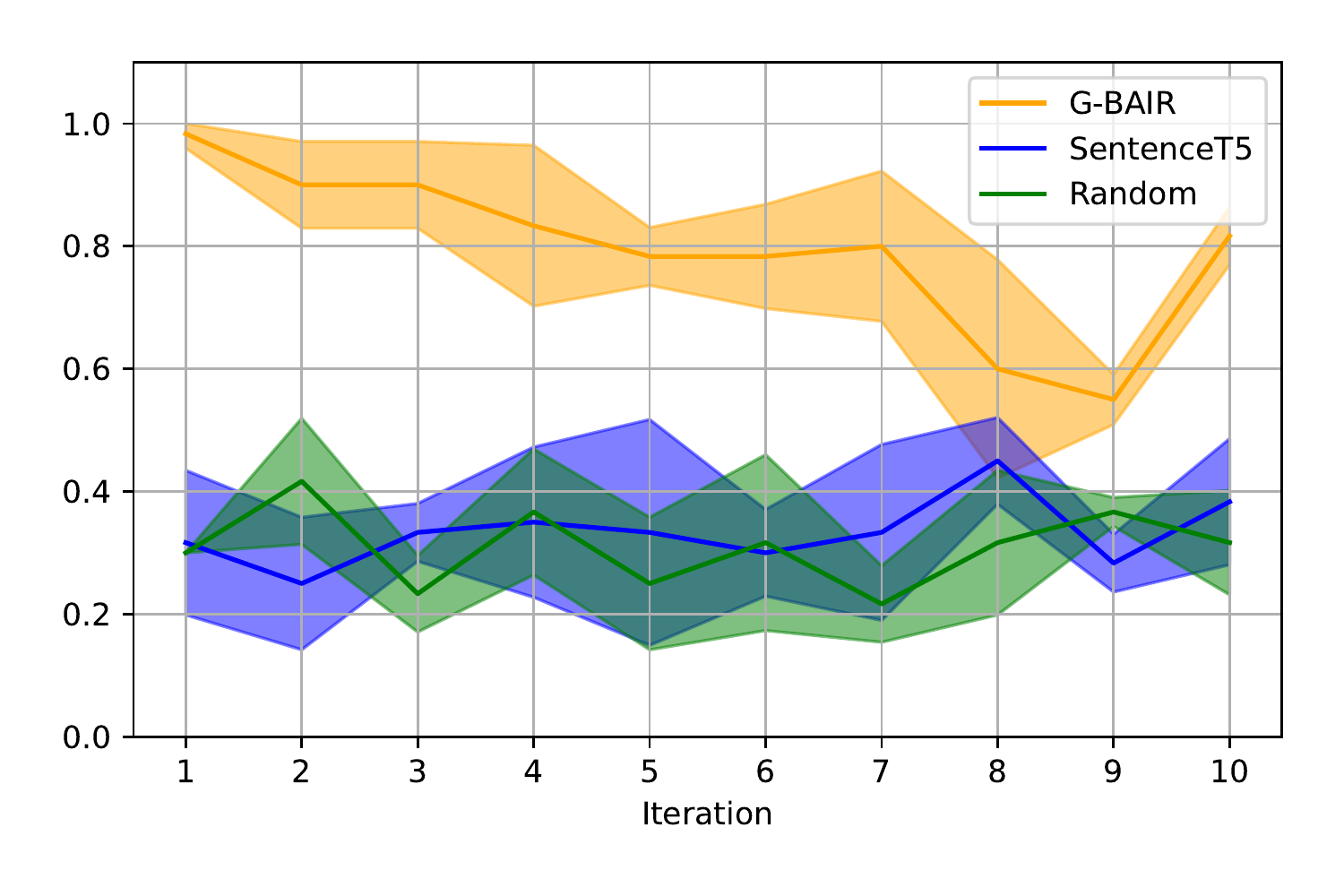}
         \caption{Fraction of corrupted examples identified by TracIn}
     \end{subfigure}
     \hfill
    \caption{Illustration of model performance recovery for \tfivexxl{} on \parlaistandard{} in terms of AP (a) and the fraction of identified corrupted examples per iteration (b). Results are averaged across three independent runs with the standard deviations shown.}
    \label{fig:tfive_xxl_standard}
\end{figure*}

\begin{figure*}[!ht]
     \centering
     \begin{subfigure}[b]{0.48\textwidth}
         \centering
         \includegraphics[width=\textwidth]{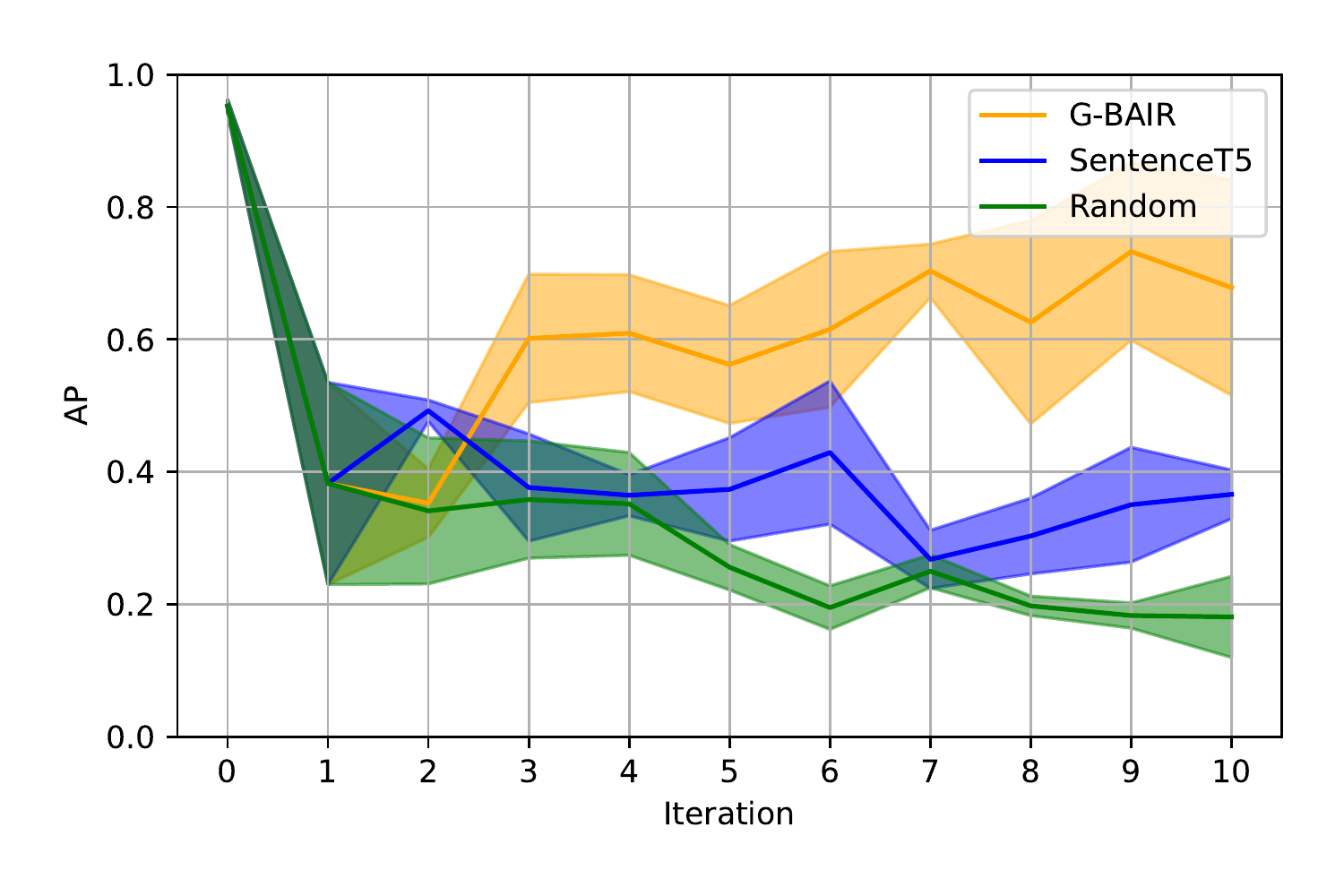}
         \caption{TracIn performance recovery}
     \end{subfigure}
     \hfill
     \begin{subfigure}[b]{0.48\textwidth}
         \centering
         \includegraphics[width=\textwidth]{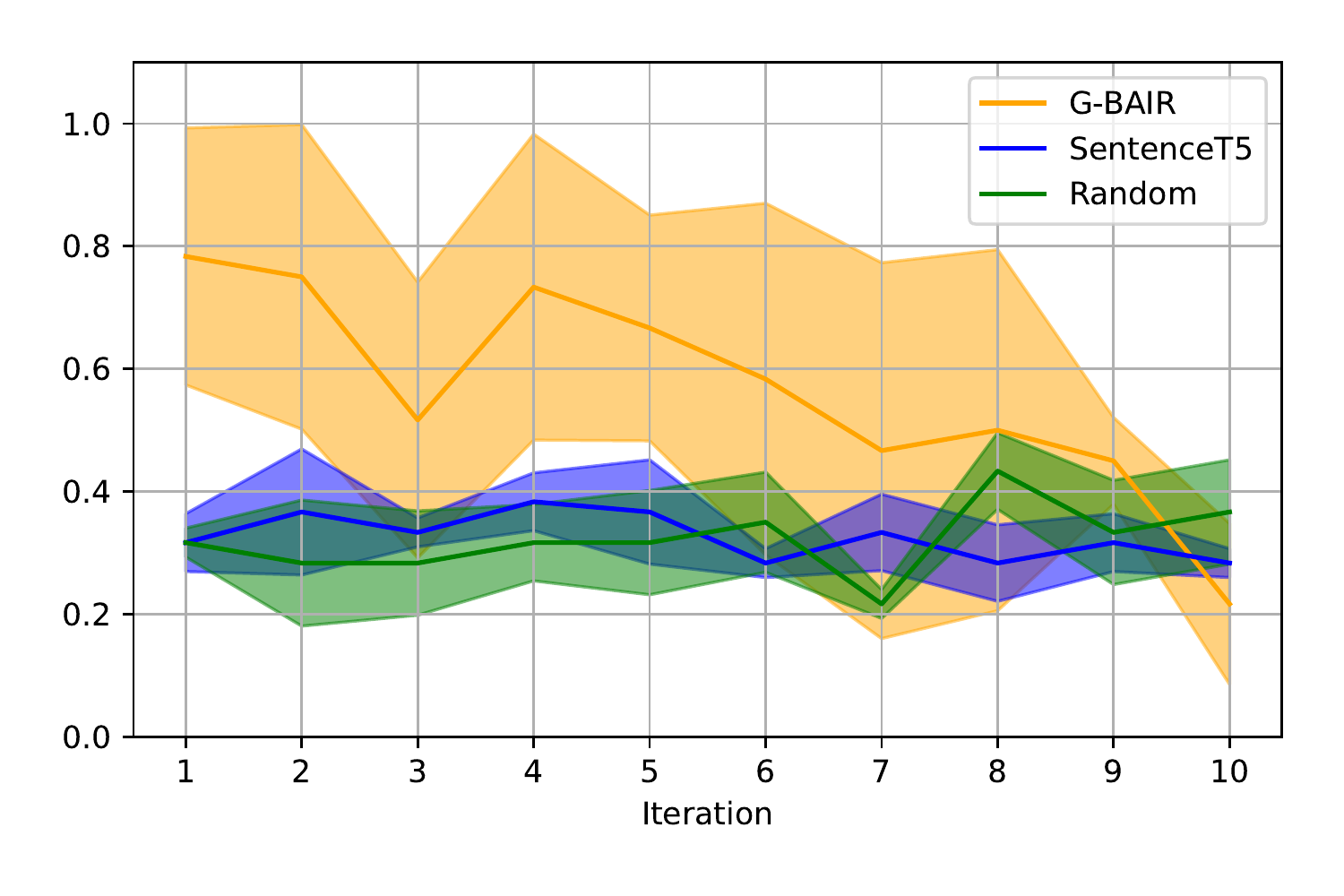}
         \caption{Fraction of corrupted examples identified by TracIn}
     \end{subfigure}
     \hfill
    \caption{Illustration of model performance recovery for \tfivexxl{} on \parlaiadversarial{} in terms of AP (a) and the fraction of identified corrupted examples per iteration (b). Results are averaged across three independent runs with the standard deviations shown.}
    \label{fig:tfive_xxl_adversarial}
\end{figure*}

\begin{figure*}[!ht]
     \centering
     \begin{subfigure}[b]{0.48\textwidth}
         \centering
         \includegraphics[width=\textwidth]{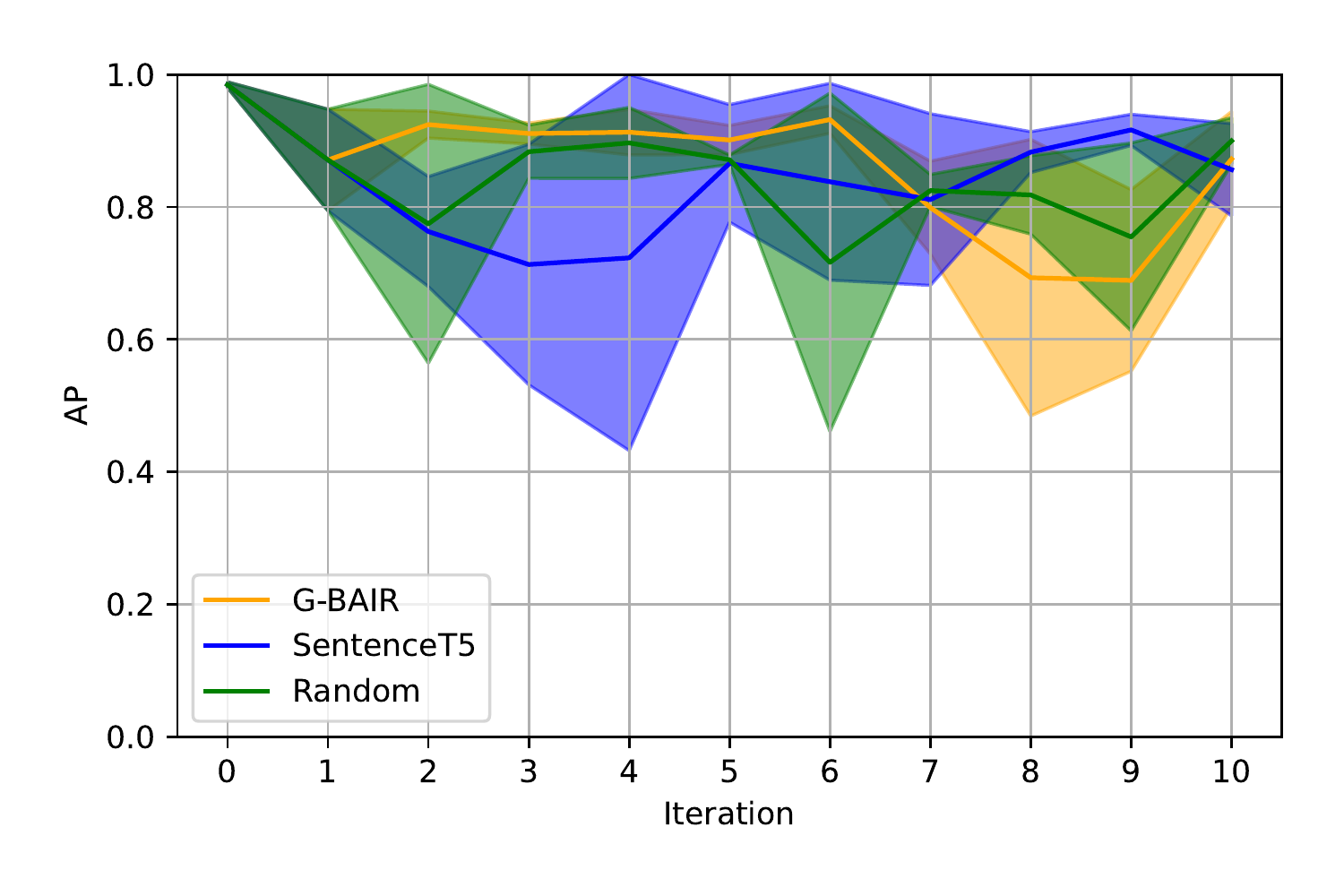}
         \caption{TracIn performance recovery}
     \end{subfigure}
     \hfill
     \begin{subfigure}[b]{0.48\textwidth}
         \centering
         \includegraphics[width=\textwidth]{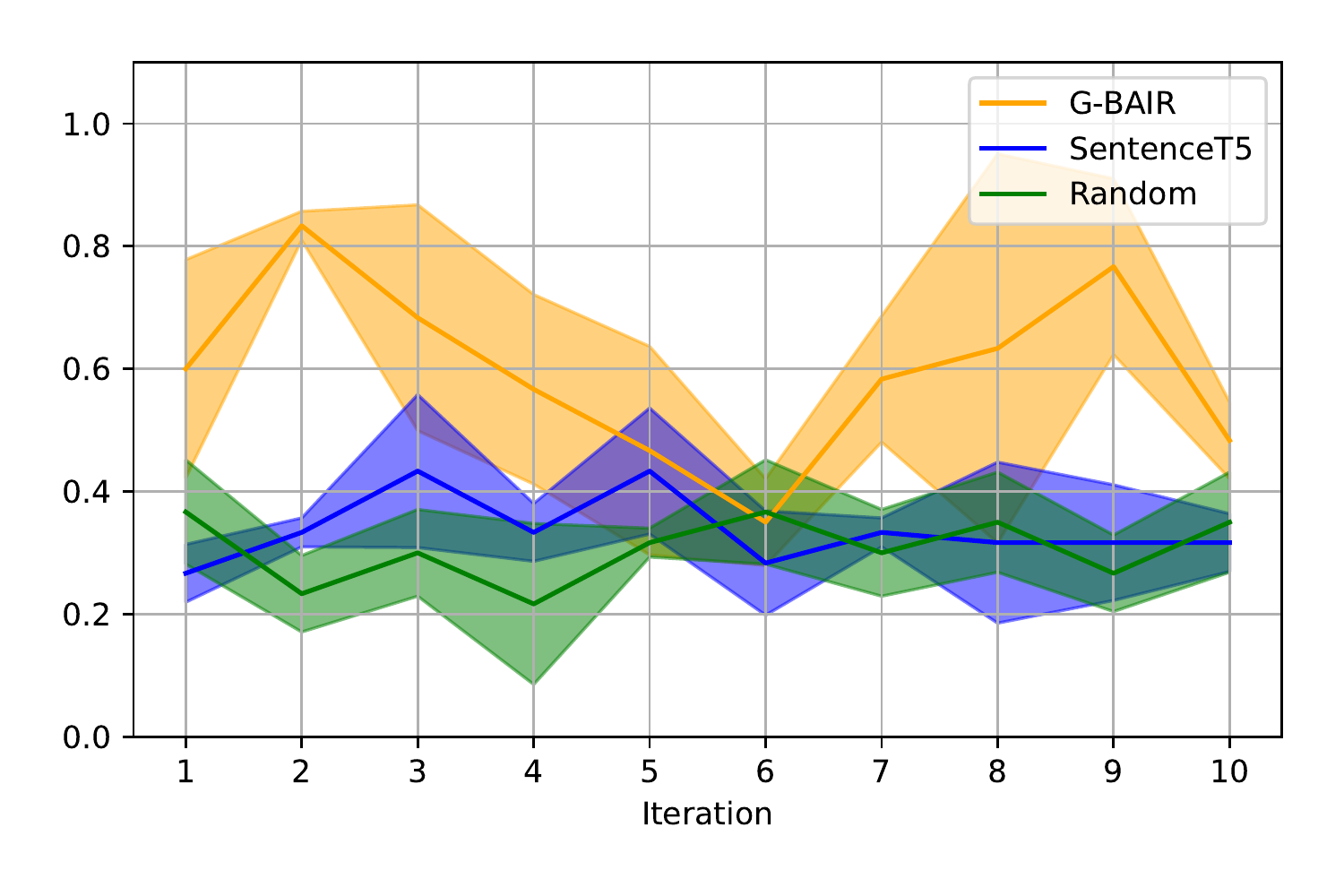}
         \caption{Fraction of corrupted examples identified by TracIn}
     \end{subfigure}
     \hfill
    \caption{Illustration of model performance recovery for \palm{} on \parlaistandard{} in terms of AP (a) and the fraction of identified corrupted examples per iteration (b). Results are averaged across three independent runs with the standard deviations shown.}
    \label{fig:palm_standard}
\end{figure*}

\begin{figure*}[!ht]
     \centering
     \begin{subfigure}[b]{0.48\textwidth}
         \centering
         \includegraphics[width=\textwidth]{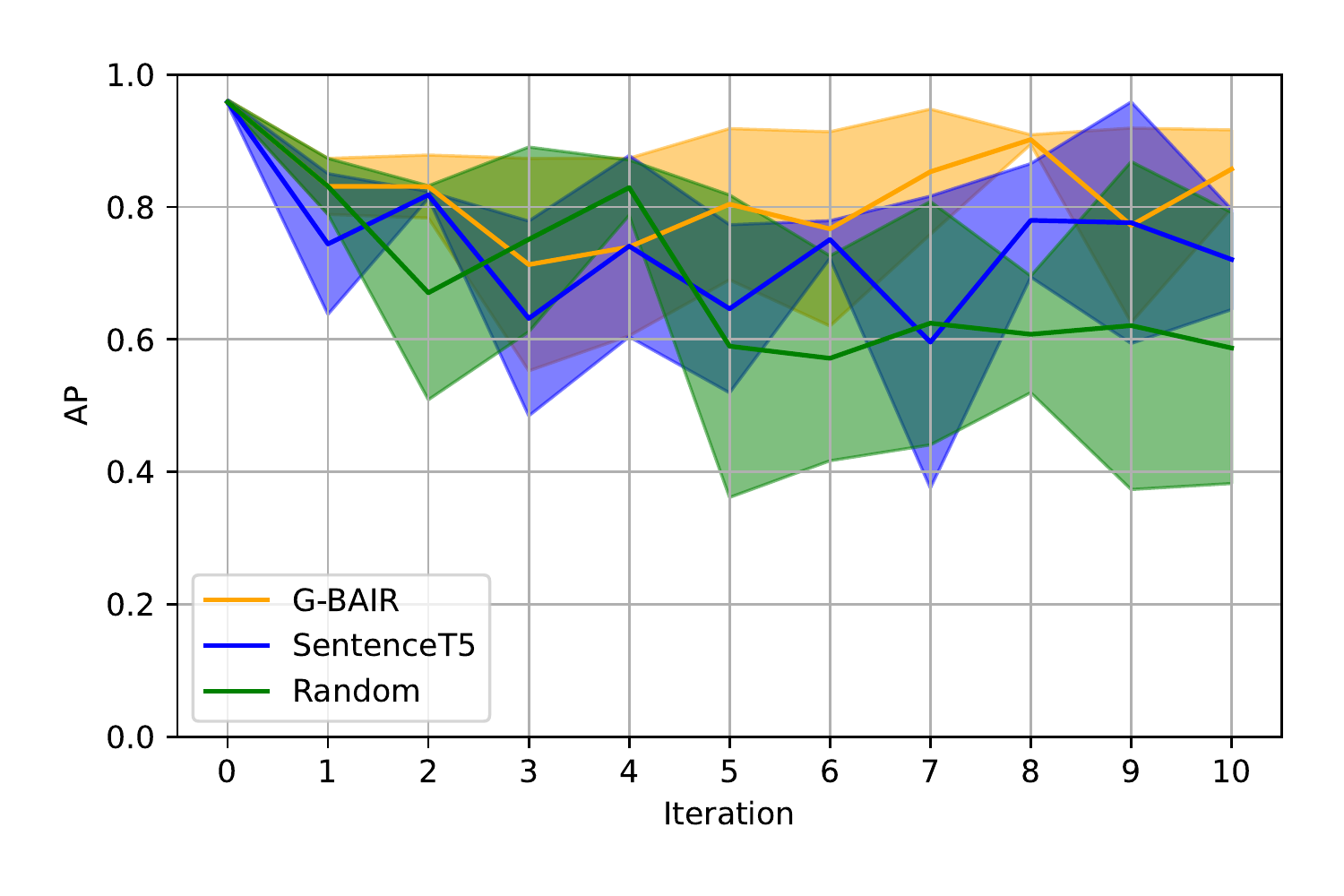}
         \caption{TracIn performance recovery}
     \end{subfigure}
     \hfill
     \begin{subfigure}[b]{0.48\textwidth}
         \centering
         \includegraphics[width=\textwidth]{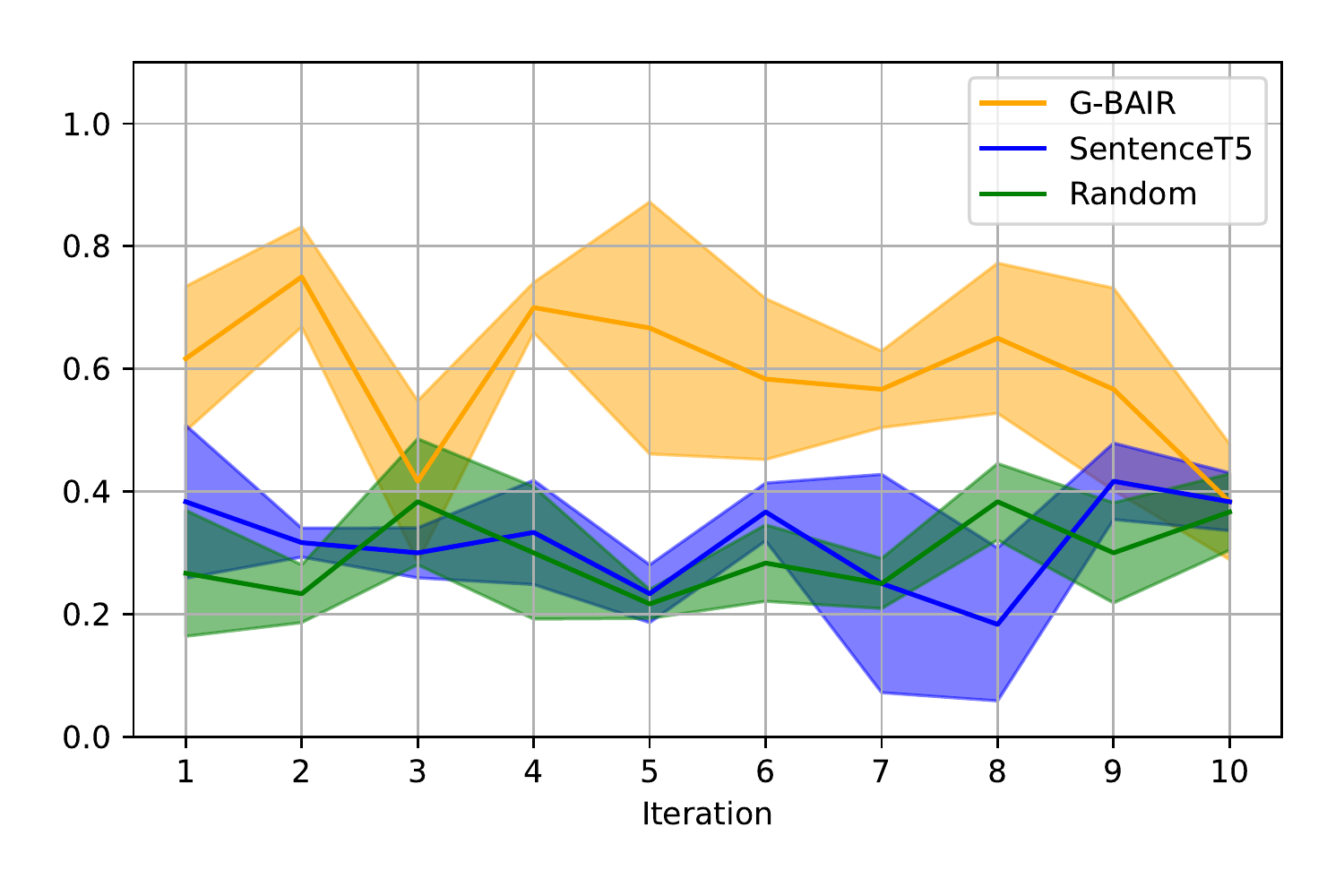}
         \caption{Fraction of corrupted examples identified by TracIn}
     \end{subfigure}
     \hfill
    \caption{Illustration of model performance recovery for \palm{} on \parlaiadversarial{} in terms of AP (a) and the fraction of identified corrupted examples per iteration (b). Results are averaged across three independent runs with the standard deviations shown.}
    \label{fig:palm_adversarial}
\end{figure*}

Additional illustrations analogous to the ones in Figure~\ref{fig:plot_tfive_base_adversarial} can found in Figure~\ref{fig:plot_tfive_base_standard} (\tfivebase{} on \parlaistandard{}), Figure~\ref{fig:tfive_xxl_standard} (\tfivexxl{} on \parlaistandard{}), Figure~\ref{fig:tfive_xxl_adversarial} (\tfivexxl{} on \parlaiadversarial{}), Figure~\ref{fig:palm_standard} (\palm{} on \parlaistandard{}), and Figure~\ref{fig:palm_adversarial} (\palm{} on \parlaiadversarial{}).

\section{Baseline results with USE}
\label{app:baseline_use}
To assess potential effects from using \sentencetfive{} as our semantic encoder for the baseline comparison, we conducted additional experiments using universal sentence encoders~\cite[\use{};][]{cer2018universal} instead of \sentencetfive{}. Performance recovery scores for these experiments can be found in Table~\ref{table:main_tracin_results_use}. In line with results in the main paper (Table~\ref{table:main_tracin_results}), we can see little to no improvement when using \use{} as vector representations for measuring data influence. For AP, \use{} performs similar to \random{} (with the exception of \tfivexxl{} on \parlaiadversarial{}) and is hence largely outperformed by \gbair{}. A similar picture emerges for CI$^2$R, where the scores for \use{} are close to the \random{} baseline across experiments, indicating that using universal sentence encodings does not aid in identifying corrupted examples more than randomly choosing examples from the training set.

\begin{table*}[!ht]
    \centering
    \resizebox{\textwidth}{!}{
    \begin{tabular}{c l c c c c c c c c}
    \toprule
    & & \multicolumn{5}{c}{\textbf{AP}} & \multicolumn{3}{c}{\textbf{CI}$^2$\textbf{R}} \\ 
    \cmidrule(lr){3-7} \cmidrule(lr){8-10}
    \textbf{Dataset} & \textbf{Model} & \textbf{Clean} & \textbf{Corrupted} & \random{} & \use{} & \gbair{} & \random{} & \use{} & \gbair{} \\
    \midrule
    \multirow{3}{*}{\parlaistandard{}} 
    & \tfivebase{} & $0.92_{0.01}$ & $0.31_{0.09}$ & $0.39_{0.05}$ & $0.39_{0.09}$ & $0.61_{0.08}$ & $0.20_{0.01}$ & $0.19_{0.03}$ & $0.43_{0.01}$ \\
    & \tfivexxl{} & $0.97_{0.00}$ & $0.45_{0.12}$ & $0.36_{0.03}$ & $0.41_{0.10}$ & $0.76_{0.11}$ & $0.21_{0.03}$ & $0.20_{0.01}$ & $0.53_{0.02}$ \\
    & \palm{} & $0.98_{0.00}$ & $0.87_{0.08}$ & $0.90_{0.04}$ & $0.92_{0.01}$ & $0.93_{0.02}$ & $0.20_{0.00}$ & $0.15_{0.02}$ & $0.40_{0.01}$ \\
    \midrule
    \multirow{3}{*}{\parlaiadversarial{}} 
    & \tfivebase{} & $0.91_{0.03}$ & $0.19_{0.03}$ & $0.27_{0.08}$ & $0.24_{0.04}$ & $0.54_{0.05}$ & $0.21_{0.03}$ & $0.24_{0.02}$ & $0.39_{0.02}$ \\
    & \tfivexxl{} & $0.95_{0.01}$ & $0.38_{0.15}$ & $0.36_{0.09}$ & $0.46_{0.04}$ & $0.73_{0.13}$ & $0.21_{0.01}$ & $0.22_{0.05}$ & $0.38_{0.14}$ \\
    & \palm{} & $0.96_{0.00}$ & $0.83_{0.04}$ & $0.83_{0.04}$ & $0.83_{0.07}$ & $0.90_{0.01}$ & $0.20_{0.01}$ & $0.18_{0.01}$ & $0.39_{0.04}$ \\
    \bottomrule
    \end{tabular}
    }
    \caption{Mean (standard deviation) performance scores in terms of average precision (AP) as well as the CI$^2$R for clean, corrupted, and recovered training sets across three seeds. For AP, \textbf{Clean} and \textbf{Corrupted} denote performances on the test set before and after corrupting 30\% of the training data. \random{} and \use{} show the recovered performances using the two baselines, and \gbair{} shows recovered performance using our proposed method.}
    \label{table:main_tracin_results_use}
\end{table*}

\end{document}